\documentclass[pmlr]{jmlr}

\usepackage{longtable}
\usepackage{booktabs}
\usepackage[nobiblatex]{xurl}
\usepackage{hhline}
\usepackage{amsfonts}
\usepackage{nicefrac}       
\usepackage{microtype}      
\usepackage{xcolor}         
\usepackage{ulem}
\usepackage{multirow}
\usepackage{adjustbox}      
\usepackage{enumitem}
\usepackage{color}
\usepackage{hyperref}

\newcommand{\eg}{{\it e.g.,~}}
\newcommand{\ie}{{\it i.e.,~}}
\newcommand{\vs}{{\it vs.~}}

\jmlrvolume{1}
\jmlryear{2022}
\jmlrworkshop{Meta-Knowledge Transfer/Communication in Different Systems}

\title[NeurIPS'22 Cross-Domain MetaDL competition]{NeurIPS'22 Cross-Domain MetaDL competition: Design and baseline results\titletag{\thanks{The authors are in alphabetical order of last name, except the first author.}}}

\author{
    \Name{Dustin Carrión-Ojeda}\textsuperscript{1} \Email{dustin.carrion@gmail.com}\\
    \Name{Hong Chen}\textsuperscript{2} \Email{h-chen20@mails.tsinghua.edu.cn}\\
    \Name{Adrian {El Baz}}\textsuperscript{3,4} \Email{eb.adrian@hotmail.fr}\\
    \Name{Sergio Escalera}\textsuperscript{5,6} \Email{sergio.escalera.guerrero@gmail.com}\\ 
    \Name{Chaoyu Guan}\textsuperscript{2} \Email{guancy19@mails.tsinghua.edu.cn} \\
    \Name{Isabelle Guyon}\textsuperscript{1,5} \Email{guyon@chalearn.org}\\
    \Name{Ihsan Ullah}\textsuperscript{1} \Email{ihsan2131@gmail.com}\\
    \Name{Xin Wang}\textsuperscript{2} \Email{xin\_wang@tsinghua.edu.cn}\\
    \Name{Wenwu Zhu}\textsuperscript{2} \Email{wwzhu@tsinghua.edu.cn}\vspace{0.3cm}\\
    {\footnotesize \normalfont \textsuperscript{1} LISN/CNRS/INRIA, Université Paris-Saclay, France\\
    \textsuperscript{2} Department of Computer Science and Technology, Tsinghua University, China\\
    \textsuperscript{3} MILA - Québec AI Institute, Montréal, Canada\\
    \textsuperscript{4} NeuroPoly Lab, Institute of Biomedical Engineering, Polytechnique Montréal, Canada \\
    \textsuperscript{5} ChaLearn, USA\\
    \textsuperscript{6} Computer Vision Center, Universitat de Barcelona, Spain}
}

\editor{Editor's name}

\begin{document}

\maketitle

\begin{abstract}
We present the design and baseline results for a new challenge in the \href{https://metalearning.chalearn.org/}{ChaLearn meta-learning series}, accepted at NeurIPS'22, focusing on ``cross-domain'' meta-learning. Meta-learning aims to leverage experience gained from previous tasks to solve new tasks efficiently (\ie with better performance, little training data, and/or modest computational resources). While previous challenges in the series focused on {\it within-domain} few-shot learning problems, with the aim of learning efficiently {\it N-way k-shot} tasks (\ie \textit{N} class classification problems with \textit{k} training examples), this competition challenges the participants to solve ``any-way'' and ``any-shot'' problems drawn from various domains (healthcare, ecology, biology, manufacturing, and others), chosen for their humanitarian and societal impact. To that end, we created \href{https://meta-album.github.io/}{Meta-Album}, a meta-dataset of 40 image classification datasets from 10 domains, from which we carve out tasks with any number of ``ways'' (within the range 2-20) and any number of ``shots'' (within the range 1-20). The competition is with code submission, fully blind-tested on the \href{https://codalab.lisn.upsaclay.fr/competitions/3627}{CodaLab challenge platform}. The code of the winners will be open-sourced, enabling the deployment of automated machine learning solutions for few-shot image classification across several domains.
\end{abstract}

\section{Introduction}\label{sec:introduction}
Challenges in machine learning have been instrumental in pushing the state-of-the-art and stimulating participants to tackle new difficult problems. Since 2015, ChaLearn has been organizing challenges in \href{https://automl.chalearn.org/} {Automated Machine Learning (AutoML)} \cite{Sun2019AutoML} and \href{https://autodl.chalearn.org/}{Automated Deep Learning (AutoDL)} \cite{liu2021tpami}, with the aim of reducing the need of human intervention in the design and implementation of machine learning models, to the greatest possible extent. Our challenge series gave rise to the popular \href{{https://automl.github.io/auto-sklearn/master/}}{auto-sklearn} software and outlined the importance of good representations (obtained from pre-trained backbone networks), data augmentation, and meta-learning. These results prompted us to organize a new \href{https://metalearning.chalearn.org/}{ChaLearn challenge series in meta-learning}, focusing first on image classification and few-shot learning. This challenge, the {\bf NeurIPS'22 Cross-Domain MetaDL}, is the third edition in the series. {\bf Submissions are open between July 1 and August 31, 2022.} The results will be presented at the NeurIPS'22 conference.

Traditionally, image classification has been tackled using deep learning methods whose performance relies on the availability of large amounts of data \cite{Phoo2021}. Recent efforts in meta-learning \cite{Jamal2019} have contributed to making a lot of progress in few-shot learning for image classification problems. {\em Tasks} or ``episodes'' are made of a certain {\em number of classes} or ``ways'' and {\em number of labeled examples per class} or ``shots''. Despite progress made, allowing the community to reach accuracies above $90\%$ in the last ChaLearn meta-learning challenge \cite{elbaz2021pmlr}, evaluation protocols have a common drawback: they focus only on within-domain few-shot learning, \ie even when evaluated on multiple domains (\eg insect classification, texture classification, satellite images, etc.), models meta-trained on a given domain are meta-tested on the {\em same domain}. As documented in the literature, within-domain few-shot learning approaches have poor generalization ability to unrelated domains \cite{Phoo2021}. Nevertheless, this kind of generalization is crucial since there are scenarios where only one or two examples per class are available (\eg rare birds or plants), and there is no close domain with enough information to be used as source domain. Therefore, addressing domain variations has become a research area of great interest. Additionally, in most works about few-shot learning, the number of ways and shots is fixed, which is not always the case in real application scenarios.

Currently, the most popular benchmark used for cross-domain meta-learning is Meta-Dataset \cite{Meta-Dataset2020}. This benchmark tackles the problems mentioned above by including 10 image classification datasets from several application domains in one collection and analyzing the impact of using a variable number of ways and shots. Although it has been widely used to evaluate state-of-the-art methods \cite{SUR2020,URT2020,FLUTE2021,URL2021,TSA2021}, it cannot be used in our competition because it is already well-known by the meta-learning community. Additionally, the datasets in Meta-Dataset have a large variance in the number of classes and examples per class, introducing bias in our competition design.

The main contributions of this paper are the design of a new challenge in the ChaLearn meta-learning series and the presentation of baseline results. Our new design will challenge participants to {\bf generalize across domains} in {\bf different regimes in numbers of ways and shots}, and {\bf compare ``de novo" training with the use of pre-trained backbones}. Note that ``de novo" training means that the algorithm needs to learn from scratch without using any previous knowledge such as pre-trained backbones, \ie the backbones must be initialized randomly. In conjunction with the organization of this challenge, we developed a large meta-dataset called \href{https://meta-album.github.io/}{Meta-Album} described in a companion paper \cite{meta-album-2022}, including 40 datasets belonging to 10 different domains, relevant to ``AI for good'', such as ecology, medicine, and biology, with the intent of maximizing the economic and societal impact of the challenge. In this competition, 30 of these datasets will be used for meta-training and meta-testing, then released publicly as a {\bf long-lasting benchmark} to further push the state-of-the-art. A single (final) submission will be evaluated during the final challenge phase, using ten datasets previously unused by the meta-learning community. The {\bf code of the winners will be open-sourced} and enable {\bf practical AutoML applications} since the meta-trained learner will be readily usable for few-shot image classification in the 10 domains of the challenge.  

\section{Problem setting} \label{sec:problem_setting}
This challenge has two motivational scenarios: (1) Few-shot image classification and (2) Meta-learning from limited amounts of meta-learning data. For the former problem, we target users wishing to create an image classifier from a few pictures of each class (\eg taken with a smartphone) in a {\bf new domain} (\eg classify clouds). The challenge winning solution(s) should make this possible for {\bf any-way any-shot} in the range [2-20] ``ways'' (classes) and [1-20] ``shots'' (training examples per class). For the latter problem, the solution of the winner should deliver a meta-learning algorithm leveraging knowledge from previous tasks, without relying on pre-trained backbones, applicable to a wider range of applications than image classification (encouraged by the prize distribution, see \appendixref{apd:leagues}). This section first explains the setting of the previous MetaDL challenge organized for NeurIPS'21 (within-domain few-shot learning). Then, it explains the new variant we developed for NeurIPS'22 (cross-domain any-way any-shot learning). Both competitions are with code submission and the participants must supply code following a designated API \cite{Liu2019}, featuring Python objects (see \appendixref{apd:api}): \texttt{MetaLearner}, \texttt{Learner}, and \texttt{Predictor}. \texttt{MetaLearner} uses meta-training data (a dataset of datasets) to create \texttt{Learner}; \texttt{Learner} in turn uses training examples (images) to return \texttt{Predictor}; finally, \texttt{Predictor} uses unlabeled test examples to return predicted class labels. The competitions are composed of 2 main phases, a {\bf feedback phase} with many submissions allowed and immediate feedback provided on a leaderboard, and a {\bf final test phase} with only 1 submission tested on new datasets. In both phases, data are not visible to the participants; only the code submitted has access to evaluation data. Ground truth labels of test data are kept secret and are only visible to the scoring program.

\subsection{Within-domain few-shot learning} \label{sec:metadl}
The NeurIPS'21 MetaDL competition focused on ``within-domain'' few-shot learning image classification in the \textit{N}-way \textit{k}-shot setting  \cite{elbaz2021metadl}. The \texttt{Learner} was meta-tested on many 5-way 5-shot tasks carved out from several multi-class image datasets, each task including $N=5$ classes drawn at random, with $k=5$ examples per class in the support (training) set and 20 examples per class in the query (test) set. Half of the classes of each dataset were reserved for meta-training and the other half for meta-testing. During meta-training, the \texttt{MetaLearner} could choose the configuration of data received: (1) batch training with examples of all classes within the domain at hand, (2) episodic training with examples grouped in tasks having a support and a query set. Importantly, meta-learning and meta-testing were performed using classes from the {\bf same dataset}, which we refer to as {\bf within domain meta-learning}. Submissions made by participants were then ranked per dataset, and the final ranking was obtained by averaging such ranks. Five datasets from 5 domains (ecology, bio-medicine, manufacturing, optical character recognition, and remote sensing) were used in the feedback phase, and 5 other fresh datasets from the same domains were used in the final evaluation phase. All datasets had at least 20 classes and 40 images per class \cite{elbaz2021pmlr}.

\subsubsection{Lessons learned and limitations} \label{sec:lessons_learned}
The NeurIPS'21 MetaDL competition considered a refined competition protocol developed for a previous MetaDL competition \cite{elbaz2021metadl}, introducing multiple domains, which added additional sophistication in terms of scoring as well as GPU-time budgeting. However, the setting remained relatively simplified since the meta-training and meta-testing were performed within-domain (\ie non-overlapping classes of the same dataset were used for meta-training and meta-testing) using meta-test tasks with a fixed number of ways and shots. The winners \cite{metadelta} obtained over 92\% accuracy on all 5 domains in the final phase (with complete blind-testing of their code). Thus, this indicates that we can move to more complicated problems. Following these observations, the Cross-Domain MetaDL challenge intends to mix tasks from multiple domains and present variable numbers of \textit{ways} and \textit{shots}.

Although the NeurIPS'21 MetaDL competition did not constrain participants to use deep-learning, {\em de facto}, all participants based their solutions on deep-learning models with convolutions (specifically, either convolutional neural networks or transformer models). Additionally, fine-tuning on meta-training data turned out to be important. However, there are indications that off-the-shelf backbones pre-trained with self-supervised learning on massive datasets might be the most promising approach, essentially making meta-learning unnecessary for image classification problems. Thus, meta-learning should be benchmarked in {\em de novo} training conditions to prepare for scenarios (in other domains) in which such backbones are not available.

As reported by several top-ranking teams, meta-learning was possible within domains (in the form of fine-tuning pre-trained backbones), but MAML-style episodic meta-learning did not turn out to be more effective than vanilla pre-training with gradient descent. Based on the embedding generated by the backbones, prototypical classifiers seem more efficient than linear classifiers. Hence, the Cross-Domain MetaDL challenge also allows further probing of the effectiveness of various meta-learning solutions.

\subsection{From within-domain to cross-domain any-way any-shot learning} \label{sec:cd-metadl}
Following the lessons learned from the NeurIPS'21 competition, the new Cross-Domain MetaDL challenge aims to push the complete automation of few-shot learning by demanding participants to design learning agents capable of producing a trained classifier in the \textbf{cross-domain any-way any-shot setting}. 

As introduced in \sectionref{sec:metadl}, the few-shot learning problems are often referred as \textit{N}-way \textit{k}-shots problems. In these problems, each task $ \mathcal{T}_j = \{ \mathcal{D}_{\mathcal{T}_j}^{train}, \mathcal{D}_{\mathcal{T}_j}^{test}\}$ consists of a small training set $\mathcal{D}_{\mathcal{T}_j}^{train}$ and a small test set $\mathcal{D}_{\mathcal{T}_j}^{test}$, referred to as \textit{support} and \textit{query} sets, respectively. The number of ways \textit{N} denotes the number of classes in a task that represents an image classification problem, the same $N$ classes are present in $\mathcal{D}_{\mathcal{T}_j}^{train}$ and $\mathcal{D}_{\mathcal{T}_j}^{test}$. The number of shots \textit{k} denotes the number of examples per class in the \textit{support set}. In this challenge, the tasks at meta-test time have a number of classes varying from 2 to 20 ($N \in [2, 20]$), the support set contains 1 to 20 labeled examples per class ($k \in [1, 20]$), and the query set contains 20 unlabeled examples per class, \ie $|\mathcal{D}_{\mathcal{T}_j}^{train}| = N \times k$, and $|\mathcal{D}_{\mathcal{T}_j}^{test}| = N \times 20$. Moreover, since in this competition, the tasks come from the \textbf{cross-domain} scenario, the data contained in one task $\mathcal{T}_j$ belongs strictly to one dataset. Nonetheless, different tasks may come from different datasets because the meta-dataset used to carved out the tasks is composed of multiple datasets, \ie $\mathcal{M}_{\mathcal{D}} = \{\mathcal{D}_1, \dots, \mathcal{D}_n\}$. The number of datasets $n$ in the meta-dataset $\mathcal{M}_{\mathcal{D}}$ depends on the phase (see \sectionref{sec:data}). 

The proposed setting consists of three stages: meta-training, meta-validation (optional), and meta-testing, which are used for meta-learning, model selection, and evaluation, respectively. During the meta-training stage, the participants can choose to use data in the form of \textit{tasks} $\mathcal{T}_j$ or \textit{batches} which are a collection of sampled examples from a single large dataset resulting of concatenating all datasets of the meta-training dataset, \ie $\mathcal{D}^{train} = concat(\mathcal{D}_1, \dots, \mathcal{D}_n)$. Additionally, they can specify their preferred configurations for the selected data format at this stage. The meta-validation stage is optional; therefore, it is up to the participants to use it, but the data for this stage is always in the form of tasks. Nevertheless, the participants can still specify their preferred configurations for the meta-validation tasks. Lastly, during the meta-testing stage, the participants have no control over the data, which always arrives in the form of \textit{any-way any-shot tasks} with $N \in [2, 20]$ and $k \in [1, 20]$. During meta-testing, the labels of the query set are hidden from the participants' codes.

\section{Competition design}

\subsection{Data} \label{sec:data}
The datasets of this competition belong to the \href{https://meta-album.github.io/}{Meta-Album meta-dataset}, prepared in conjunction with this competition \cite{meta-album-2022}. It consists of 40 re-purposed or novel image datasets from 10 domains: small and large animals, plants and plant diseases, vehicles, human actions, microscopic data, satellite images, industrial textures, and printed characters. We preprocessed data in a \href{https://github.com/ihsaan-ullah/meta-album/tree/master/DataFormat}{standard format} suitable for few-shot learning. The \href{https://github.com/ihsaan-ullah/meta-album/tree/master/PreProcessing}{preprocessing pipeline} includes image resizing with anti-aliasing filters into a uniform shape of 128x128x3 pixels. For this competition, we selected 30 datasets from the meta-dataset and partitioned them into 3 sets of 10 datasets, one from each domain, used in the various competition phases (Set-0, Set-1, and Set-2). All final test phase datasets are novel to the meta-learning community (not part of past meta-learning benchmarks). Sets 0-2 will be released on OpenML \cite{openml} after the competition ends.

\subsection{Competition protocol} \label{sec:protocol}
NeurIPS'22 Cross-Domain MetaDL is an online competition with code submission, \ie the participants need to provide their solutions as raw Python code that will be executed on our dedicated CodaLab site\footnote{\label{competition-site}CodaLab site for the Cross-Domain MetaDL competition: \url{https://codalab.lisn.upsaclay.fr/competitions/3627}}. Detailed competition rules are found in \appendixref{apd:rules}. To guarantee fairness in the evaluation of the participants, the CodaLab server used in this challenge is equipped with 10 identical computer workers. Each has the following configuration: 4 CPU cores, 1 Tesla T4 GPU, 16GB RAM, and 120GB storage. 

The competition follows the problem setting described in \sectionref{sec:cd-metadl}. It is composed of 3 phases. During the \textbf{Public phase} (June 15-30, 2022), no submissions can be made; instead, the participants can use the tutorial provided as part of the starting kit (see \appendixref{apd:important-links}) and Set-0 to test their solutions on their computers or \href{https://colab.research.google.com/drive/1ek519iShqp27hW3xtRiIxmrqYgNNImun?usp=sharing}{Google Colab}. Then, during the \textbf{Feedback phase} (July 1 - August 31, 2022), participants can make 2 submissions per day and a maximum of 100 submissions during the whole phase. Each submission is evaluated on 1000 any-way any-shot tasks carved out from Set-1 (100 tasks per dataset) different from the ones used for meta-training (Public data). Additionally, each submission cannot take more than 5 hours of running time. Lastly, during the \textbf{Final phase} (September 1-30, 2022), the last submission of each participant on the \textbf{Feedback phase}, whose performance is above the baseline performance (see \sectionref{sec:metric}), will be evaluated on 6000 any-way any-shot tasks carved out from Set-2 (600 tasks per dataset). Due to the increment of meta-test tasks, the allowed running time will increase to 9 hours. 

The submissions must follow our defined API (see \appendixref{apd:api}), which was designed to be flexible enough to allow participants to explore any type of meta-learning algorithms. To encourage a diversity of participants and types of submissions, the Cross-Domain MetaDL competition has 5 different leagues. \appendixref{apd:leagues} details the leagues and prizes. Notably, there is a league to {\bf encourage meta-training from scratch (``de novo" training)} as opposed to using pre-trained backbones.

\subsection{Challenge metrics} \label{sec:metric}
Since the meta-test tasks have different configurations in the number of ways and shots, this competition uses the balanced classification accuracy (bac) as the evaluation metric, normalized with respect to the number of ways (which is the number of classes in the task). This metric is defined as follows:
\begin{equation} \label{eq:metric}
    \textrm{Normalized Accuracy} = \frac{bac - bac_{RG}}{1-bac_{RG}},
\end{equation}
where $bac$, also known as the macro-averaging recall, is defined as:
\begin{equation}
    bac = \frac{1}{num\_ways} \sum_{i=1}^{num\_ways}{\frac{\textrm{correctly classified examples of class } i}{\textrm{total examples of class } i}},
\end{equation}
and $bac_{RG}$ is the accuracy of random guessing, \ie $\frac{1}{num\_ways}$. Note that by (\ref{eq:metric}), a normalized accuracy of 0 means that the performance of the submission is equivalent to random guessing. Moreover, the normalized accuracy can be negative, indicating that the submission is worse than random guessing, and the maximum achievable normalized accuracy is 1.

The error bars correspond to 95\% confidence intervals of the mean normalized accuracy at task level computed as follows:
\begin{equation}
    CI = \pm t \times \frac{\sigma}{\sqrt{n}}, 
\end{equation}
where $t$ is the \textit{t-value} depending on confidence level and degrees of freedom ($df = n-1$); $\sigma$ corresponds to the standard deviation of the normalized accuracy obtained on all meta-test tasks, and $n$ is the number of such tasks. 

In this competition, CI calculations are only indicative and not used to select winners or declare ties. The baseline performance the participants in the Feedback phase must surpass to enter into the Final phase depends on the league. The baseline performance for the Free-style and Meta-learning leagues is 0.587 and 0.361, respectively (see \appendixref{apd:leagues} for the definition of leagues). These baseline performances were calculated by averaging the normalized accuracy achieved by the best methods (see \sectionref{sec:results}) in each league over 10 runs varying the random seed of the baseline methods. 

To select the winners in the Final phase, all eligible entries are run three times, with various random seeds. The average normalized classification accuracy over all meta-test tasks is computed in each run, and the lowest of the three runs is used for the final ranking. Ties are broken according to the first submission made. Note that the baseline performances quoted in \sectionref{sec:results} are obtained by averaging the performance over multiple runs to reduce variance, while the final evaluation of participants is made based on the worst performance over three runs.

\section{Baseline results}
In this section, we present experiments to evaluate the difficulty of the new challenge setting. We run several baseline methods to evaluate whether: (1) ``Cross-domain meta-learning any-way any-shot'' (new setting) is significantly more complicated than ``Within domain 5-way 5-shot'' (old setting); (2) Baseline methods perform significantly better when using a backbone pre-trained on ImageNet rather than meta-training (or training) ``from scratch''; (3) the choice of datasets is appropriate to separate method performances.

\subsection{Baseline methods}
This competition provides six baseline methods as part of its ``starting kit''. The first one, \textbf{Train-from-scratch}, does not perform any meta-training; instead, it directly learns each meta-testing task using only its support set. The second one, \textbf{Fine-tuning}, is a simple transfer learning method consisting of pre-training a backbone network with batches of data from the concatenated meta-training datasets and then only fine-tuning the last layer at meta-test time. Three of the remaining baselines are popular meta-learning methods: \textbf{Matching Networks} \cite{matchingnetwork}, \textbf{Prototypical Networks} \cite{Snell2017}, and \textbf{FO-MAML} \cite{Finn2017}. Furthermore, the last baseline is an adaptation of MetaDelta++ \cite{metadelta}, which corresponds to the solution of the winners of the NeurIPS'21 challenge. All the baseline methods were carefully selected, aiming to have a variety of approaches in terms of training strategy (batch and episodic training) and modeling choices (fine-tuning, metric-based, and ensemble). A detailed description of each method is presented in \appendixref{apd:hyperparameters}. All baseline methods but MetaDelta++ use a ResNet-18 backbone \cite{resnet} with the best-reported hyperparameters by the original authors on 5-way 5-shot miniImageNet (see \appendixref{apd:hyperparameters}). For all baselines, the backbone can be either initialized with random weights or weights pre-trained on ImageNet (as in Meta-learning league and Free-style leagues, respectively, see \appendixref{apd:leagues} for league definition).  

\subsection{Experimental setting}

Our experiments aim to compare and contrast the protocol of the NeurIPS'21 challenge (within-domain MetaDL) with that of the NeurIPS'22 challenge (cross-domain MetaDL).

{\bf Data:} We report results for Feedback phase data of the Cross-Domain MetaDL challenge. Accordingly (see \sectionref{sec:protocol}), the meta-training and meta-testing datasets correspond to Meta-Album Set-0 and Set-1, respectively. The 10 datasets of Set-0 were divided into 7 for meta-training and 3 for meta-validation. This division was randomly made; hence, it was different in each run because of the random seed variation. For the within-domain protocol, only Set-1 was used. In this case, each dataset of Set-1 was divided into meta-training, meta-validation, and meta-testing sets with non-overlapping classes using 70\%, 15\%, and 15\% of the available classes, respectively.

{\bf Cross-Domain setting (NeurIPS'22).} The meta-learning methods were meta-trained on 30,000 5-way 10-shot tasks, the Fine-tuning baseline was meta-trained on 30,000 batches of size 16, and the MetaDelta++ baseline was meta-trained during 3.5 hours with batches of size 64. The performance of the Learners produced during the meta-training phase was validated after every 5,000 meta-training tasks (or batches in the case of the Fine-tuning method) on 300 5-way 5-shot tasks drawn from the meta-validation split except for MetaDelta++, in which case, the Learner was validated after every 50 meta-training batches on 50 5-way 5-shot tasks drawn from the meta-validation split. The query set for every task contained 20 examples per class except for the meta-validation tasks used by MetaDelta++, which contained 5 examples per class. The Learner with the best validation performance was evaluated following the protocol of the Feedback phase described in \sectionref{sec:protocol}.

{\bf Within-Domain setting (NeurIPS'21).}  We evaluated the same baseline methods in the same way as for the Cross-Domain setting but with the protocol of the NeurIPS'21 MetaDL competition. However, since in the cross-domain setting the meta-training and meta-validation sets were composed of 7 and 3 datasets, respectively, and in the within-domain setting, 1 dataset is divided into meta-training, meta-validation, and meta-testing; we adapt the number of meta-training and meta-validation iterations to have comparable results between these two protocols. Thus, the meta-learning methods were meta-trained on 4,290 5-way 10-shot tasks, the Fine-tuning baseline was pre-trained on 4,290 batches of size 16, and the MetaDelta++ baseline was pre-trained for 30 minutes with batches of size 64. The performance of the Learners produced during the meta-training phase was validated after every 750 meta-training tasks (or batches in the case of the Fine-tuning method) on 100 5-way 5-shot tasks drawn from the meta-validation split except for MetaDelta++, in which case the Learner was validated after every 50 meta-training batches on 50 5-way 5-shot tasks drawn from the meta-validation split. To resemble the NeurIPS'21, during meta-testing, the configuration for the tasks was 5-way 5-shot.

{\bf Computational resources.} All experiments are carried out with the same resources as used in the Cross-Domain MetaDL competition (see \sectionref{sec:protocol}).

\begin{figure}[t]
    \floatconts
    {fig:baselines}
    {\caption{{\bf Comparison of ``within-domain'' and ``cross-domain'' few-shot learning using a randomly initialized backbone and a pre-trained backbone.} Both barplots show the average normalized accuracy over 3,000 meta-test tasks (100 tasks per dataset in each run). In addition, the meta-test task configuration on the left barplot is 5-way 5-shot while on the right is any-way any-shot. The corresponding 95\% CIs are computed at task level.}}
    {
        \subfigure[Within-Domain]{
          \includegraphics[width=0.46\linewidth]{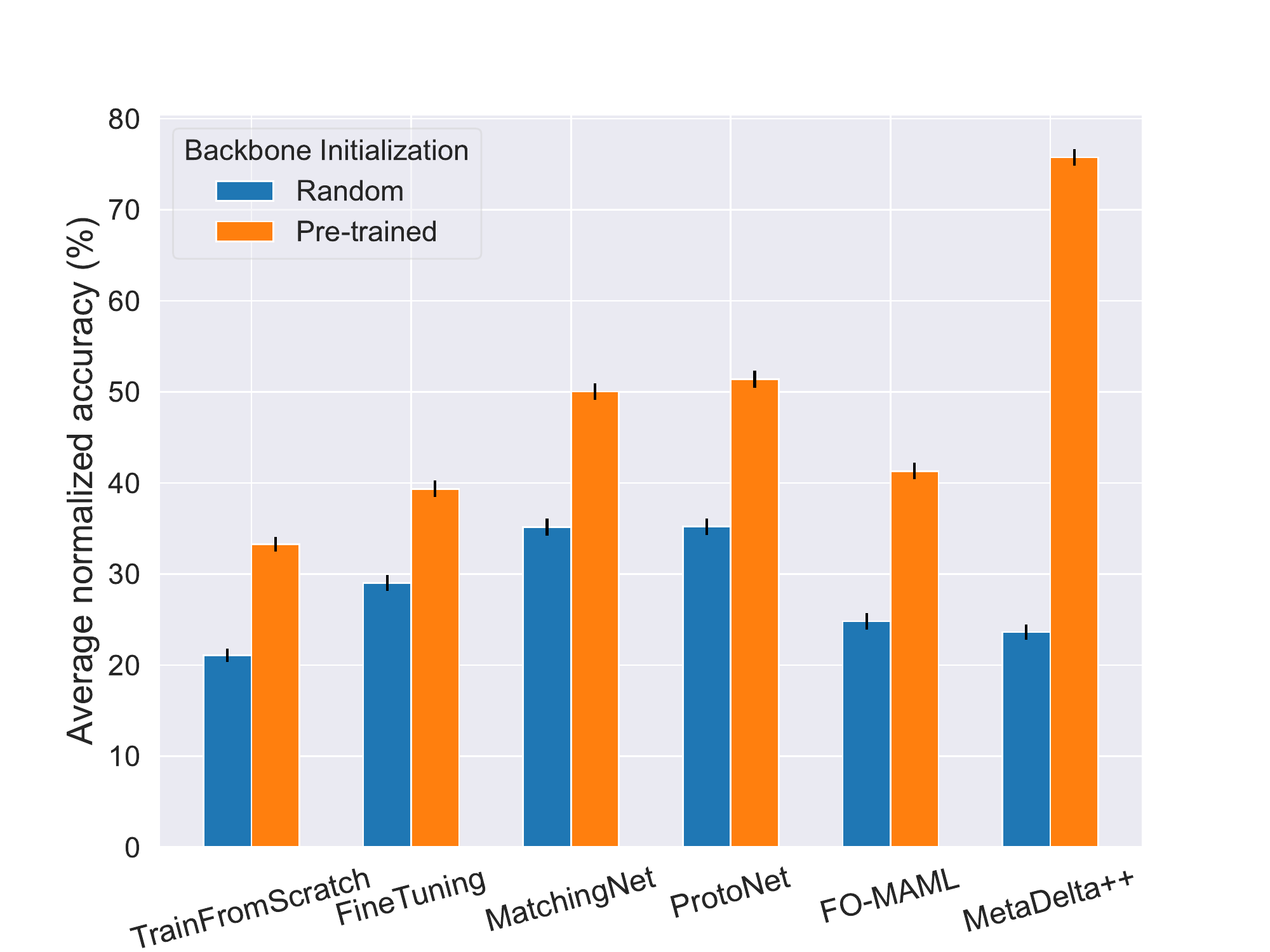}}
        \qquad
        \subfigure[Cross-Domain]{
          \includegraphics[width=0.46\linewidth]{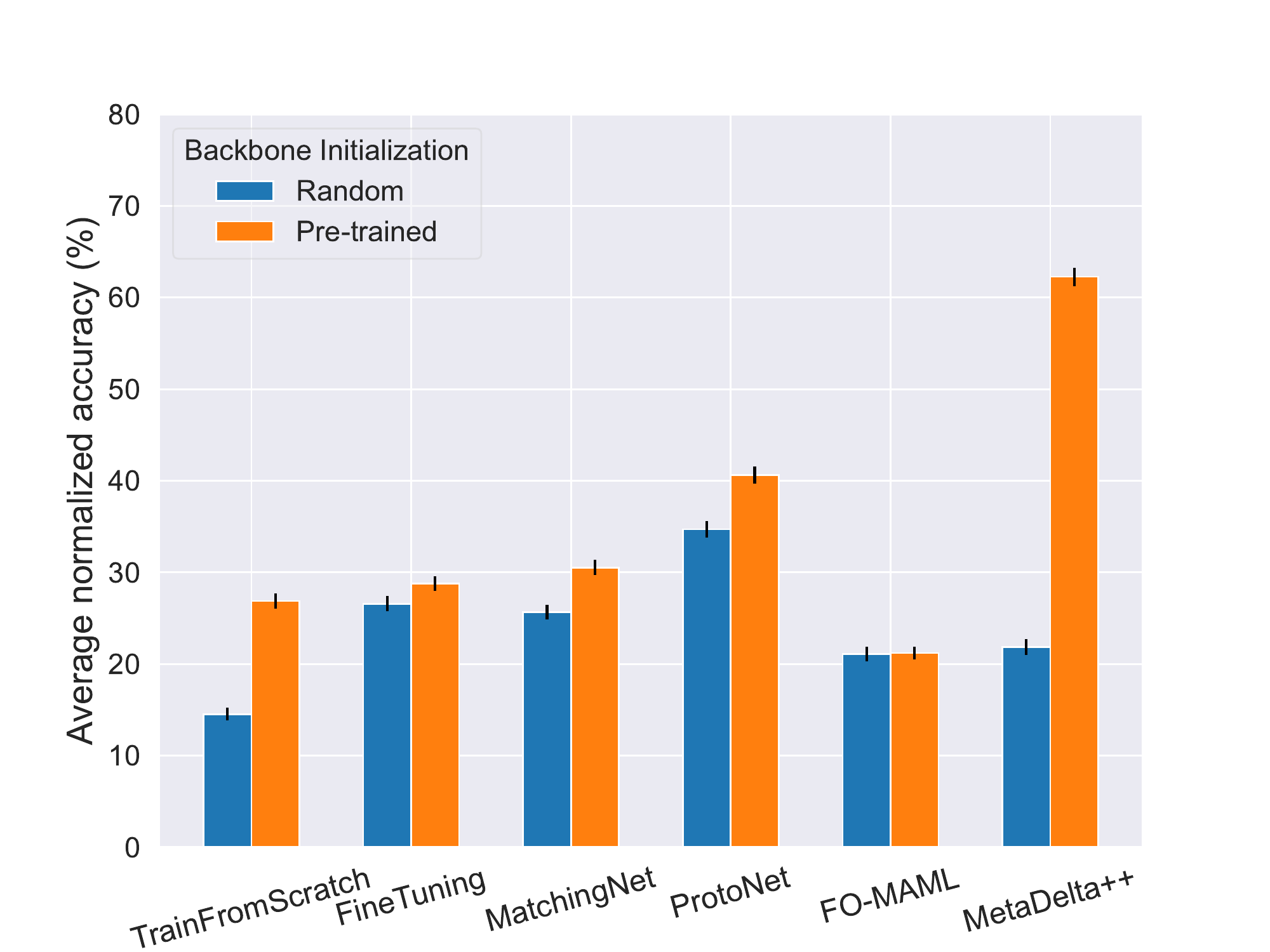}}
      }
\end{figure}

\subsection{Results}\label{sec:results}
We aggregated results in various manners to compare the settings of the NeurIPS'21 and the NeurIPS'22 challenges with respect to (1) within-domain \vs cross-domain and (2) pre-trained \vs randomly initialized backbone.

\begin{figure}[t]
    \floatconts
    {fig:config-impact}
    {\caption{{\bf Comparison of the influence of the number of ways and shots on the performance in the ``cross-domain'' setting using a pre-trained backbone.} We plot the average normalized accuracy achieved by the baselines using pre-trained weights. The corresponding 95\% CIs are computed at task level.}}
    {
        \subfigure[Influence of the number of ways]{
          \includegraphics[width=0.46\linewidth]{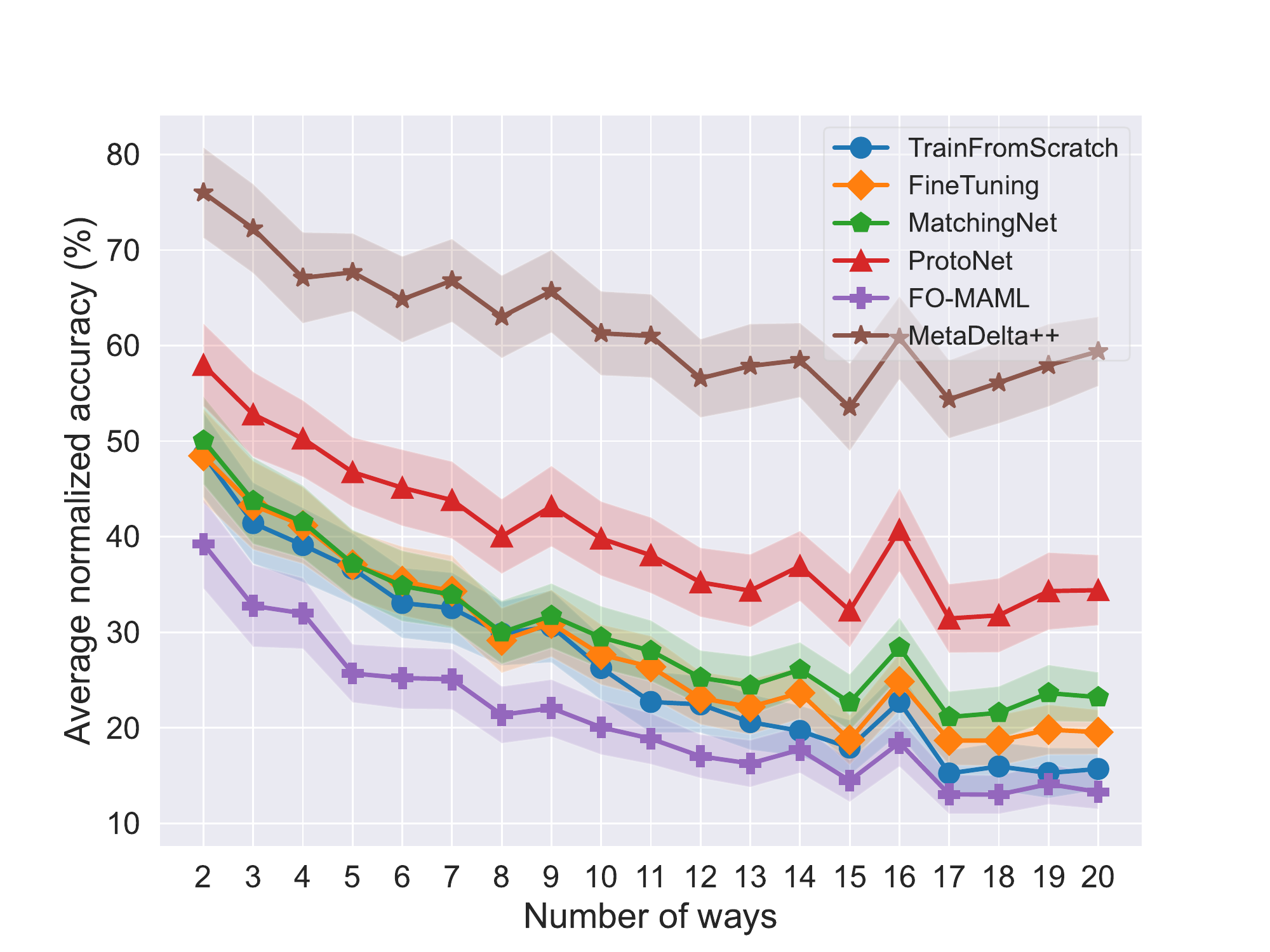}}
        \qquad
        \subfigure[Influence of the number of shots]{
          \includegraphics[width=0.46\linewidth]{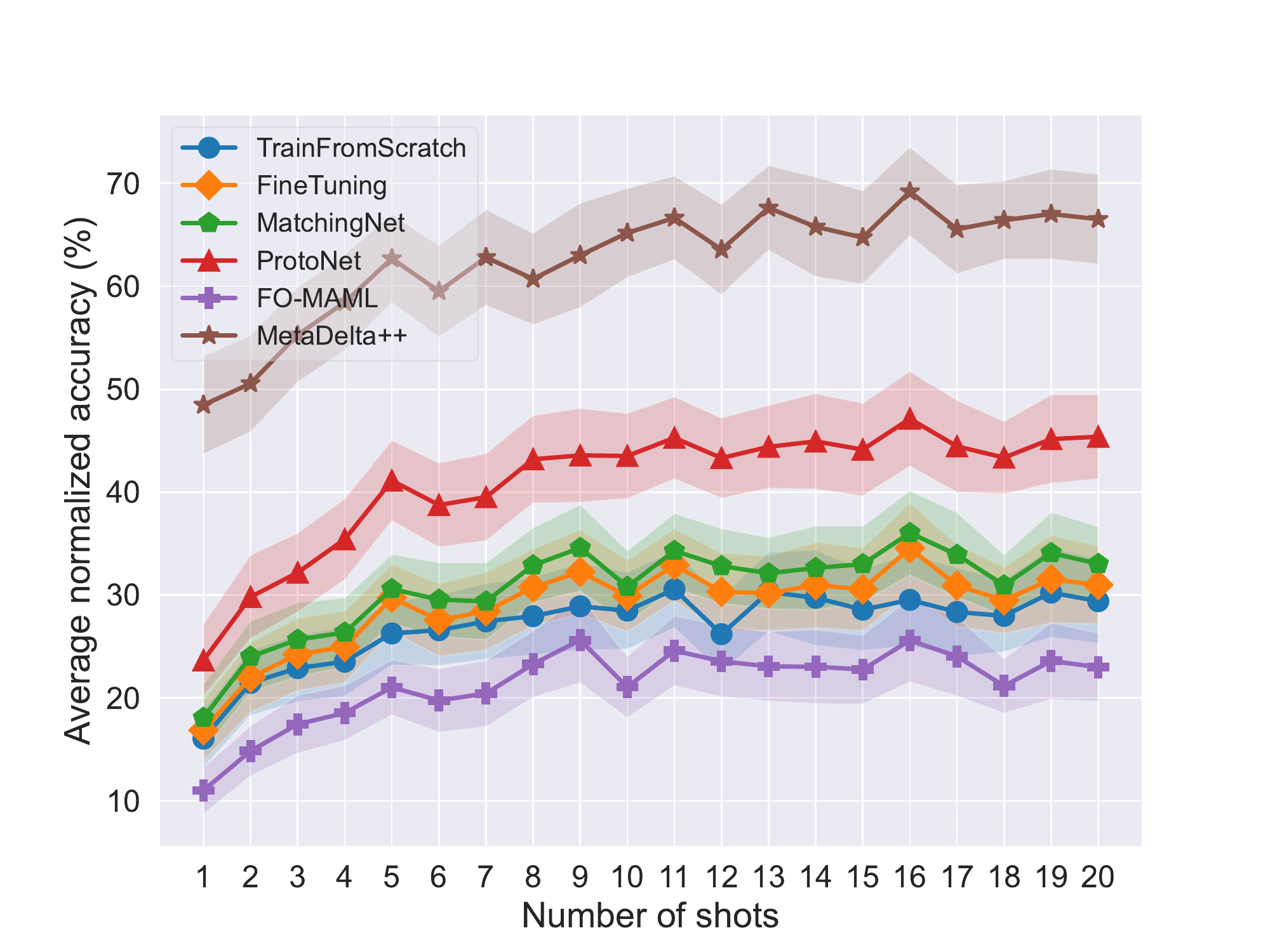}}
      }
\end{figure}

{\bf Method comparison.} In \figureref{fig:baselines}, we compare baseline methods by averaging results over all tasks from all datasets. Before meta-training, the backbone networks are initialized with random or pre-trained weights on ImageNet. The figure shows that initializing the backbones with pre-trained weights helps significantly, indicating that perhaps our meta-training set is not large enough or that the meta-training time is insufficient. We hope to see improvements in the Meta-learning league of the challenge regarding using random initialization. Moreover, the winner of the previous challenge (MetaDelta++) performs significantly better than other baselines when using a pre-trained backbone. However, Prototypical Networks is the best option when no pre-training is allowed. Additionally, we see that the new cross-domain setting is more complicated than the within-domain setting. In \figureref{fig:config-impact}, we study the influence of the number of ways and shots on the method performance in the new ``cross-domain'' setting. To that end, we averaged results over tasks with the same configuration (number of ways and shots) from all datasets and plotted the normalized accuracy. Notably, the curves do not cross, indicating that {\bf the ranking of methods is not influenced by the number of ways and shots}. We show only results using pre-trained backbone networks because the curves obtained with randomly initialized weights are qualitatively similar (only worse, and ordered differently, as in \figureref{fig:baselines}). As expected, performances degrade with the number of ways and increase with the number of shots. Interestingly, the most significant increment occurs up to 5 shots. \appendixref{apd:results} contains the detailed results for all figures presented in this section.

\begin{figure}[t]
    \floatconts
    {fig:datasets-difficulty}
    {\caption{{\bf Difficulty comparison of feed-back data} for ``within-domain'' and ``cross-domain'' few-shot learning, with a randomly initialized or pre-trained backbone. The top of the blue bar indicates the worst baseline performance (Train-from-scratch without pre-training). The top of the orange bar indicates the best baseline performance (MetaDelta++ with pre-training). The top of the green bar indicates the maximum achievable performance. The larger the green bar, the larger the {\it intrinsic difficulty}. The larger the orange bar, the larger the {\it modeling difficulty}. The average normalized accuracy was computed over 300 meta-test tasks (100 tasks per dataset in each run). Left: 5-way 5-shot; Right: any-way any-shot.}}
    {
        \subfigure[Within-Domain]{
          \includegraphics[width=0.46\linewidth]{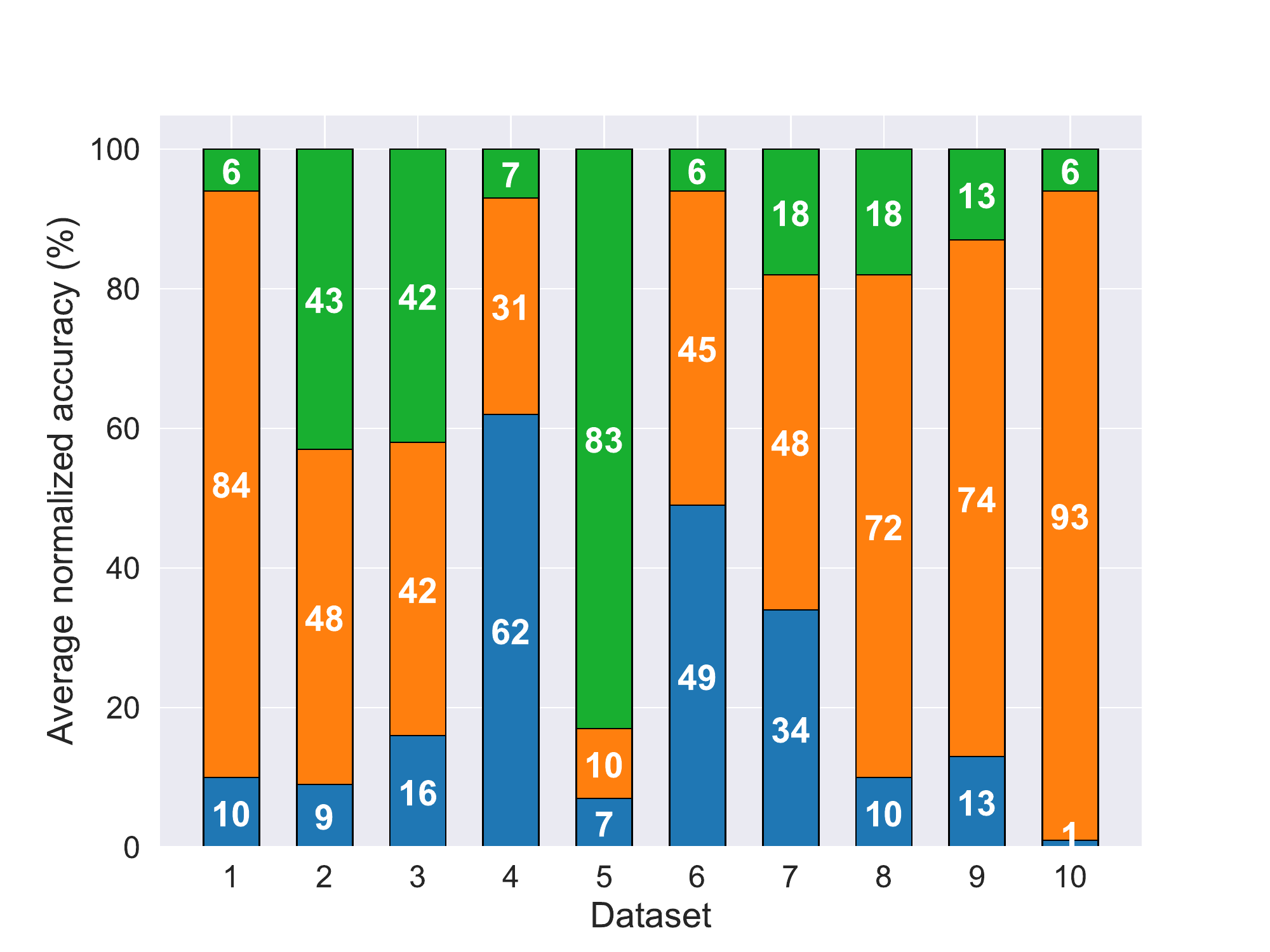}}
        \qquad
        \subfigure[Cross-Domain]{
          \includegraphics[width=0.46\linewidth]{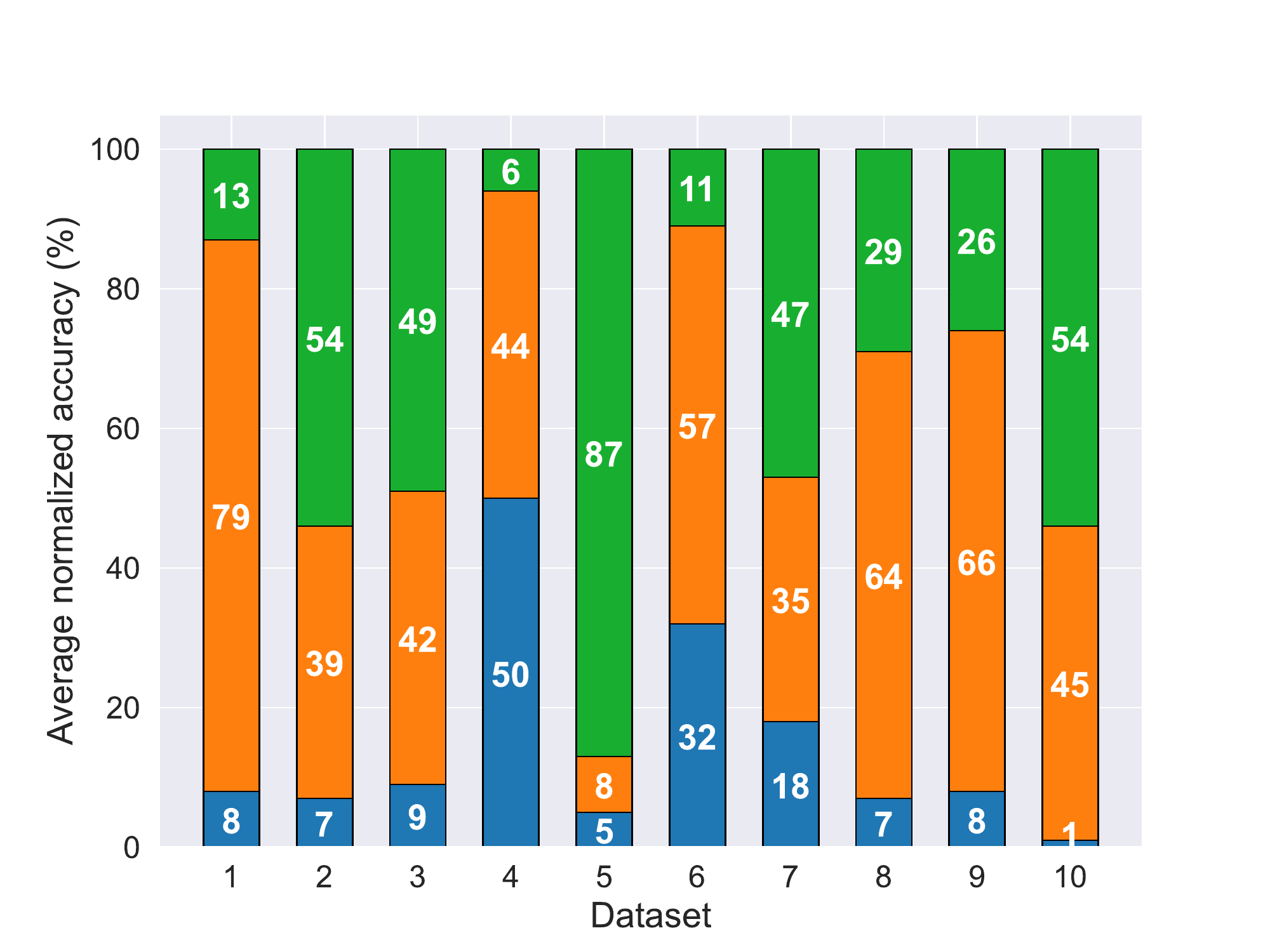}}
      }
\end{figure}

{\bf Dataset comparison.} In \figureref{fig:datasets-difficulty}, we averaged performances per dataset and reported only the results of the worst baseline (Train-from-scratch without pre-training) and the best baseline (MetaDelta++ with pre-training, previous challenge winners). These performances allow us to evaluate the intrinsic difficulty of the datasets (difference between the maximum achievable performance and the performance of the best baseline method -- green bar) and the modeling difficulty (difference between the best and worst baseline methods -- orange bar). As can be seen, the datasets show a range of difficulty, from dataset 4, which seems relatively easy, even to the worst baseline, to dataset 5, which is challenging even for the best baseline. Most datasets (except 5) have a reasonably large orange bar, indicating that the performance of methods spread over an extensive range, which is desirable in a challenge to separate methods. Dataset 10 is an interesting case: the best method performed well in the within-domain setting, but its performance dropped significantly in the cross-domain setting. We find that this domain does not resemble others; hence this is not so surprising that meta-learning within the domain should be more favorable. Generally, performances drop when we move to the new cross-domain setting; thus, the participants of the new challenge have some margin for improvement. 

\section{Conclusion and further work}
We evaluated several baselines covering a variety of approaches to tackle few-shot learning problems to compare the protocols of NeurIPS'21 and NeurIPS'22 challenge settings. The experimental results show that the new proposed any-way any-shot cross-domain setting is more challenging than the previously studied 5-way 5-shot within-domain setting. This increment in problem complexity will allow us to encourage the participants to aim at finding methods capable of learning from multiple domains and generalize to all those domains in a more realistic test environment. Additionally, our findings show that if pre-trained backbones are allowed, MetaDelta++ is the best option among the baselines. In general, all baselines (except for FO-MAML) benefit from using pre-trained initialization. However, if using pre-trained weights is not allowed, which is the case for some real-world applications where no pre-trained backbone is available, Prototypical Networks is the best option within the evaluated methods. Moreover, our experiments allowed us to estimate the difficulty level of each dataset used in the Feedback phase of the new Cross-Domain MetaDL competition. The observed modeling difficulty is a good motivation for this competition since there is room for improvement, which is the expected outcome of this challenge. Finally, these results show that, due to the differences among domains, the difficulty of some datasets increases significantly in the new setting compared to the previous one.

While this competition studies cross-domain meta-generalization across 10 domains, it does not challenge participants to meta-generalize out of these domains since meta-test data includes new datasets from these exact 10 domains. We plan to organize a ``domain independent'' sequel, in which datasets from new domains not seen during meta-training will be used for meta-testing.

\acks{We acknowledge support from ChaLearn, ANR AI chair HUMANIA ANR-19-CHIA-0022, TAILOR, an ICT48 network funded by EU Horizon 2020 program GA 952215, and the help of Mike Huisman to create the baselines code (except MetaDelta++), and of Romain Mussard, Manh Hung Nguyen, and Gabriel Lauzzana who worked as competition beta-testers. Experiments were performed using a Google cloud grant.}

\bibliography{references}

\begin{thebibliography}{20}
\providecommand{\natexlab}[1]{#1}
\providecommand{\url}[1]{\texttt{#1}}
\expandafter\ifx\csname urlstyle\endcsname\relax
  \providecommand{\doi}[1]{doi: #1}\else
  \providecommand{\doi}{doi: \begingroup \urlstyle{rm}\Url}\fi

\bibitem[Chen et~al.(2021)Chen, Guan, Wei, Wang, and Zhu]{metadelta}
Yudong Chen, Chaoyu Guan, Zhikun Wei, Xin Wang, and Wenwu Zhu.
\newblock {MetaDelta: A Meta-Learning System for Few-shot Image
  Classification}.
\newblock In I~Guyon, J.~N. van Rijn, S.~Treguer, and J.~Vanschoren, editors,
  \emph{Proceedings of the AAAI Workshop on Meta-Learning and MetaDL
  Challenge}, volume 140 of \emph{Proceedings of Machine Learning Research},
  pages 17--28. PMLR, 2021.
\newblock URL \url{https://proceedings.mlr.press/v140/chen21a.html}.

\bibitem[Dvornik et~al.(2020)Dvornik, Schmid, and Mairal]{SUR2020}
Nikita Dvornik, Cordelia Schmid, and Julien Mairal.
\newblock {Selecting Relevant Features from a Universal Representation for
  Few-shot Classification}.
\newblock \emph{arXiv preprint}, 2020.
\newblock URL \url{https://arxiv.org/abs/2003.09338}.

\bibitem[El~Baz et~al.(2021)El~Baz, Guyon, Liu, van Rijn, Treguer, and
  Vanschoren]{elbaz2021metadl}
Adrian El~Baz, Isabelle Guyon, Zhengying Liu, Jan~N. van Rijn, Sebastien
  Treguer, and Joaquin Vanschoren.
\newblock {Advances in MetaDL: AAAI 2021 challenge and workshop}.
\newblock In I~Guyon, J.~N. van Rijn, S.~Treguer, and J.~Vanschoren, editors,
  \emph{Proceedings of the AAAI Workshop on Meta-Learning and MetaDL
  Challenge}, volume 140 of \emph{Proceedings of Machine Learning Research},
  pages 1--16. PMLR, 2021.
\newblock URL \url{https://proceedings.mlr.press/v140/el-baz21a.html}.

\bibitem[El~Baz et~al.(2022)El~Baz, Ullah, Alcobaça, Carvalho, Chen, Ferreira,
  Gouk, Guan, Guyon, Hospedales, Hu, Huisman, Hutter, Liu, Mohr, Öztürk, van
  Rijn, Sun, Wang, and Zhu]{elbaz2021pmlr}
Adrian El~Baz, Ihsan Ullah, Edesio Alcobaça, André C. P. L.~F. Carvalho, Hong
  Chen, Fabio Ferreira, Henry Gouk, Chaoyu Guan, Isabelle Guyon, Timothy
  Hospedales, Shell Hu, Mike Huisman, Frank Hutter, Zhengying Liu, Felix Mohr,
  Ekrem Öztürk, Jan~N. van Rijn, Haozhe Sun, Xin Wang, and Wenwu Zhu.
\newblock {Lessons learned from the NeurIPS 2021 MetaDL challenge: Backbone
  fine-tuning without episodic meta-learning dominates for few-shot learning
  image classification}.
\newblock In D.~Kiela, M.~Ciccone, and B.~Caputo, editors, \emph{Proceedings of
  the NeurIPS 2021 Competitions and Demonstrations Track}, volume 176 of
  \emph{Proceedings of Machine Learning Research}, pages 80--96. PMLR, 2022.
\newblock URL \url{https://proceedings.mlr.press/v176/el-baz22a.html}.

\bibitem[Finn et~al.(2017)Finn, Abbeel, and Levine]{Finn2017}
Chelsea Finn, Pieter Abbeel, and Sergey Levine.
\newblock {Model-Agnostic Meta-Learning for Fast Adaptation of Deep Networks}.
\newblock In D.~Precup and Y.~W. Teh, editors, \emph{Proceedings of the 34th
  International Conference on Machine Learning}, volume~70 of \emph{Proceedings
  of Machine Learning Research}, pages 1126--–1135. PMLR, 2017.
\newblock URL \url{https://proceedings.mlr.press/v70/finn17a.html}.

\bibitem[Guyon et~al.(2019)Guyon, Sun{-}Hosoya, Boull{\'{e}}, Escalante,
  Escalera, Liu, Jajetic, Ray, Saeed, Sebag, Statnikov, Tu, and
  Viegas]{Sun2019AutoML}
Isabelle Guyon, Lisheng Sun{-}Hosoya, Marc Boull{\'{e}}, Hugo~Jair Escalante,
  Sergio Escalera, Zhengying Liu, Damir Jajetic, Bisakha Ray, Mehreen Saeed,
  Mich{\`{e}}le Sebag, Alexander~R. Statnikov, Wei{-}Wei Tu, and Evelyne
  Viegas.
\newblock {Analysis of the AutoML Challenge Series 2015--2018}.
\newblock In F.~Hutter, L.~Kotthoff, and J.~Vanschoren, editors,
  \emph{Automated Machine Learning - Methods, Systems, Challenges}, pages
  177--219. Springer International Publishing, 2019.
\newblock \doi{10.1007/978-3-030-05318-5_10}.

\bibitem[He et~al.(2016)He, Zhang, Ren, and Sun]{resnet}
Kaiming He, Xiangyu Zhang, Shaoqing Ren, and Jian Sun.
\newblock {Deep Residual Learning for Image Recognition}.
\newblock In \emph{Proceedings of the IEEE Conference on Computer Vision and
  Pattern Recognition (CVPR)}, pages 770--778, 2016.
\newblock \doi{10.1109/CVPR.2016.90}.

\bibitem[Jamal and Qi(2019)]{Jamal2019}
Muhammad~Abdullah Jamal and Guo-Jun Qi.
\newblock {Task Agnostic Meta-Learning for Few-Shot Learning}.
\newblock In \emph{Proceedings of the IEEE/CVF Conference on Computer Vision
  and Pattern Recognition (CVPR)}, pages 11711--11719, 2019.
\newblock \doi{10.1109/CVPR.2019.01199}.

\bibitem[Li et~al.(2021)Li, Liu, and Bilen]{URL2021}
Wei-Hong Li, Xialei Liu, and Hakan Bilen.
\newblock {Universal Representation Learning from Multiple Domains for Few-shot
  Classification}.
\newblock In \emph{Proceedings of the IEEE/CVF International Conference on
  Computer Vision (ICCV)}, pages 9506--—9515, 2021.
\newblock \doi{10.1109/ICCV48922.2021.00939}.

\bibitem[Li et~al.(2022)Li, Liu, and Bilen]{TSA2021}
Wei-Hong Li, Xialei Liu, and Hakan Bilen.
\newblock {Cross-domain Few-shot Learning with Task-specific Adapters}.
\newblock \emph{arXiv preprint}, 2022.
\newblock URL \url{https://arxiv.org/abs/2107.00358}.

\bibitem[Liu et~al.(2021{\natexlab{a}})Liu, Hamilton, Long, Jiang, and
  Larochelle]{URT2020}
Lu~Liu, William~L. Hamilton, Guodong Long, Jing Jiang, and Hugo Larochelle.
\newblock {A Universal Representation Transformer Layer for Few-Shot Image
  Classification}.
\newblock In \emph{Proceedings of the 9th International Conference on Learning
  Representations (ICLR)}, 2021{\natexlab{a}}.
\newblock URL \url{https://openreview.net/forum?id=04cII6MumYV}.

\bibitem[Liu et~al.(2019)Liu, Xu, Madadi, Jacques~Junior, Escalera, Rajaa, and
  Guyon]{Liu2019}
Zhengying Liu, Zhen Xu, Meysam Madadi, Julio Jacques~Junior, Sergio Escalera,
  Shangeth Rajaa, and Isabelle Guyon.
\newblock {Overview and unifying conceptualization of Automated Machine
  Learning}.
\newblock In \emph{Proceedings of the Automating Data Science workshop at ECML
  PKDD}, 2019.

\bibitem[Liu et~al.(2021{\natexlab{b}})Liu, Pavao, Xu, Escalera, Ferreira,
  Guyon, Hong, Hutter, Ji, Junior, Li, Lindauer, Luo, Madadi, Nierhoff, Niu,
  Pan, Stoll, Treguer, Wang, Wang, Wu, Xiong, Zela, and Zhang]{liu2021tpami}
Zhengying Liu, Adrien Pavao, Zhen Xu, Sergio Escalera, Fabio Ferreira, Isabelle
  Guyon, Sirui Hong, Frank Hutter, Rongrong Ji, Julio C. S.~Jacques Junior,
  Ge~Li, Marius Lindauer, Zhipeng Luo, Meysam Madadi, Thomas Nierhoff, Kangning
  Niu, Chunguang Pan, Danny Stoll, Sebastien Treguer, Jin Wang, Peng Wang,
  Chenglin Wu, Youcheng Xiong, Arbër Zela, and Yang Zhang.
\newblock {Winning Solutions and Post-Challenge Analyses of the ChaLearn AutoDL
  Challenge 2019}.
\newblock \emph{IEEE Transactions on Pattern Analysis and Machine
  Intelligence}, 43\penalty0 (9):\penalty0 3108--3125, 2021{\natexlab{b}}.
\newblock \doi{10.1109/TPAMI.2021.3075372}.

\bibitem[Phoo and Hariharan(2021)]{Phoo2021}
Cheng~Perng Phoo and Bharath Hariharan.
\newblock {Self-training For Few-shot Transfer Across Extreme Task
  Differences}.
\newblock In \emph{Proceedings of the 9th International Conference on Learning
  Representations (ICLR)}, 2021.
\newblock URL \url{https://openreview.net/forum?id=O3Y56aqpChA}.

\bibitem[Snell et~al.(2017)Snell, Swersky, and Zemel]{Snell2017}
Jake Snell, Kevin Swersky, and Richard Zemel.
\newblock {Prototypical Networks for Few-shot Learning}.
\newblock In I.~Guyon, U.~Von Luxburg, S.~Bengio, H.~Wallach, R.~Fergus,
  S.~Vishwanathan, and R.~Garnett, editors, \emph{Advances in Neural
  Information Processing Systems}, volume~30, 2017.
\newblock URL
  \url{https://proceedings.neurips.cc/paper/2017/file/cb8da6767461f2812ae4290eac7cbc42-Paper.pdf}.

\bibitem[Triantafillou et~al.(2020)Triantafillou, Zhu, Dumoulin, Lamblin, Evci,
  Xu, Goroshin, Gelada, Swersky, Manzagol, and Larochelle]{Meta-Dataset2020}
Eleni Triantafillou, Tyler Zhu, Vincent Dumoulin, Pascal Lamblin, Utku Evci,
  Kelvin Xu, Ross Goroshin, Carles Gelada, Kevin Swersky, Pierre-Antoine
  Manzagol, and Hugo Larochelle.
\newblock {Meta-Dataset: A Dataset of Datasets for Learning to Learn from Few
  Examples}.
\newblock In \emph{Proceedings of the 8th International Conference on Learning
  Representations (ICLR)}, 2020.
\newblock URL \url{https://openreview.net/forum?id=rkgAGAVKPr}.

\bibitem[Triantafillou et~al.(2021)Triantafillou, Larochelle, Zemel, and
  Dumoulin]{FLUTE2021}
Eleni Triantafillou, Hugo Larochelle, Richard Zemel, and Vincent Dumoulin.
\newblock {Learning a Universal Template for Few-shot Dataset Generalization}.
\newblock In M.~Meila and T.~Zhang, editors, \emph{Proceedings of the 38th
  International Conference on Machine Learning}, volume 139 of
  \emph{Proceedings of Machine Learning Research}, pages 10424--10433. PMLR,
  2021.
\newblock URL \url{https://proceedings.mlr.press/v139/triantafillou21a.html}.

\bibitem[Ullah et~al.(2022)Ullah, Carrion, Escalera, Guyon, Huisman, Mohr, van
  Rijn, Sun, Vanschoren, and Vu]{meta-album-2022}
Ihsan Ullah, Dustin Carrion, Sergio Escalera, Isabelle Guyon, Mike Huisman,
  Felix Mohr, Jan~N. van Rijn, Haozhe Sun, Joaquin Vanschoren, and Phan~Anh Vu.
\newblock {Meta-Album: Multi-domain Meta-Dataset for Few-Shot Image
  Classification}.
\newblock In \emph{Submitted to: Proceedings of the Neural Information
  Processing Systems Track on Datasets and Benchmarks}, 2022.
\newblock URL \url{https://meta-album.github.io/}.

\bibitem[Vanschoren et~al.(2014)Vanschoren, van Rijn, Bischl, and
  Torgo]{openml}
Joaquin Vanschoren, Jan~N. van Rijn, Bernd Bischl, and Luis Torgo.
\newblock {OpenML: networked science in machine learning}.
\newblock \emph{ACM SIGKDD Explorations Newsletter}, 15\penalty0 (2):\penalty0
  49--60, 2014.
\newblock \doi{10.1145/2641190.2641198}.

\bibitem[Vinyals et~al.(2016)Vinyals, Blundell, Lillicrap, kavukcuoglu, and
  Wierstra]{matchingnetwork}
Oriol Vinyals, Charles Blundell, Timothy Lillicrap, koray kavukcuoglu, and Daan
  Wierstra.
\newblock {Matching Networks for One Shot Learning}.
\newblock In D.~Lee, M.~Sugiyama, U.~Luxburg, I.~Guyon, and R.~Garnett,
  editors, \emph{Advances in Neural Information Processing Systems}, volume~29,
  2016.
\newblock URL
  \url{https://proceedings.neurips.cc/paper/2016/file/90e1357833654983612fb05e3ec9148c-Paper.pdf}.

\end{thebibliography}

\newpage
\appendix

\section{Competition Rules} \label{apd:rules}
\begin{itemize}[leftmargin=*]
\setlength{\itemsep}{0ex}
    \item \textbf{General Terms}: This challenge is governed by the \href{http://www.causality.inf.ethz.ch/GeneralChalearnContestRuleTerms.html}{General ChaLearn Contest Rule Terms}, the \href{https://github.com/codalab/codalab-competitions/wiki/Privacy}{CodaLab Terms and Conditions}, and the specific rules set forth.
    \item \textbf{Announcements}: To receive announcements and be informed of any change in rules, the participants must provide a valid email.
    \item \textbf{Conditions of participation}: Participation requires complying with the rules of the challenge. Prize eligibility is restricted by US government export regulations, see the \href{http://www.causality.inf.ethz.ch/GeneralChalearnContestRuleTerms.html}{General ChaLearn Contest Rule Terms}. The organizers, sponsors, their students, close family members (parents, sibling, spouse or children) and household members, as well as any person having had access to the truth values or to any information about the data or the challenge design giving him (or her) an unfair advantage, are excluded from participation. A disqualified person may submit one or several entries in the challenge and request to have them evaluated, provided that they notify the organizers of their conflict of interest. If a disqualified person submits an entry, this entry will not be part of the final ranking and does not qualify for prizes. The participants should be aware that ChaLearn and the organizers reserve the right to evaluate for scientific purposes any entry made in the challenge, whether or not it qualifies for prizes.
    \item \textbf{Dissemination}: The challenge is part of the official selection of the NeurIPS 2022 conference. There will be publication opportunities for competition reports co-authored by organizers and top-ranking participants.
    \item \textbf{Registration}: The participants must register to CodaLab and provide a valid email address. Teams must register only once and provide a group email, which is forwarded to all team members. Teams or solo participants registering multiple times to gain an advantage in the competition may be disqualified.
    \item \textbf{Anonymity}: The participants who do not present their results at the workshop can elect to remain anonymous by using a pseudonym. Their results will be published on the leaderboard under that pseudonym, and their real name will remain confidential. However, the participants must disclose their real identity to the organizers to claim any prize they might win. See our \href{http://www.chalearn.org/privacy.html}{privacy policy} for details.
    \item \textbf{Submission method}: The results must be submitted through this CodaLab competition site. The number of submissions per day and maximum total computational time are restrained and subject to change, according to the number of participants. Using multiple accounts to increase the number of submissions in NOT permitted, except that participants that want to enter the 2 leagues ``free-style" and ``meta-learning" are allowed to create a second account under the name ``originalID\_2" , where originalID is their other account they use to make submissions (we will ask in the fact sheets which is which). In case of problem, send email to \href{mailto:metalearningchallenge@googlegroups.com}{metalearningchallenge@googlegroups.com}. The entries must be formatted as specified on the Instructions page.
    \item \textbf{Reproducibility}: The participant should make efforts to guarantee the reproducibility of their method, which in particular implies that they should use everywhere in their code the same random seed, as exemplified in the starting kit. In the Final Phase, all submissions will be run three times with various random seeds, and the worst performance will be used for final ranking. The participants will be given 2 weeks when results are released to scrutinize the evaluation procedure and the code of the winners.
    \item \textbf{Prizes}: The three top ranking participants in each league in the Final phase (blind testing) may qualify for prizes. The last valid submission in Feedback Phase will be automatically submitted to the Final Phase for final evaluation. The participant must fill out a fact sheet briefly describing their methods. There is no other publication requirement. The winners will be required to make their code publicly available under an \href{https://opensource.org/licenses}{OSI-approved license} such as, for instance, Apache 2.0, MIT or BSD-like license, if they accept their prize, within a week of the deadline for submitting the final results. Entries exceeding the time budget will not qualify for prizes. In case of a tie, the prize will go to the participant who submitted his/her entry first. Non winners or entrants who decline their prize retain all their rights on their entries and are not obliged to publicly release their code.
\end{itemize}

\textbf{Discussion:}  The rules have been designed with the criteria of \textit{inclusiveness for all participants} and \textit{openness of results} in mind. We aim to achieve inclusiveness for all participants by allowing them to enter anonymously and providing them cycles of computation (for the feedback and final phases) on our compute resources. This way, participants that do not have ample computing resources will not be limited by this and have a fair chance to win the challenge. We aim to achieve openness of results by requiring all participants to upload their base code and, afterward, fill in a fact sheet about the used methods. The information from the fact sheets will allow us to conduct post-challenge analyzes of the winners' methods. 

\textbf{Cheating prevention:} We will execute the submissions on our compute cluster to prevent participants from cheating, and the testing datasets will remain hidden in the CodaLab platform. Peeking at the final evaluation datasets will be impossible since those datasets are not even installed on the server during the feedback phase. Improperly using the data during the final phase will be prevented by never revealing the true test labels to the Learners but only showing them to the scoring program on the platform. Moreover, different sets of datasets (Set 0-2) are used in each phase to avoid domain-specific cheating and overfitting. We will also monitor submissions and reach out to participants with suspicious submission patterns. Finally, the winners will have to open-source their code to claim their prize. All other participants will individually scrutinize their code before they earn their prize.

\newpage
\section{Starting kit and important links} \label{apd:important-links}
The starting kit for this competition contains all the baseline methods described in this paper. Additionally, since anyone interested in meta-learning can participate, we also provide a tutorial with three difficulty levels:
\begin{itemize}
    \item Beginner level (no prerequisites)
    \item Intermediate level (some knowledge of Python and meta-learning)
    \item Advanced level (solid knowledge of Python and meta-learning)
\end{itemize}
Each level includes information from previous levels. The idea of the tutorial is to explain to the participants all the necessary details about the competition, the data, and the submissions. To facilitate the usage of the starting kit, we distribute it in three different ways:
\begin{itemize}
    \item Google Colab: \url{https://colab.research.google.com/drive/1ek519iShqp27hW3xtRiIxmrqYgNNImun?usp=sharing}
    \item GitHub Repository: \url{https://github.com/DustinCarrion/cd-metadl}
    \item Zip file: \url{https://codalab.lisn.upsaclay.fr/my/datasets/download/a3476a08-a190-4455-adc2-3db175799c98}
\end{itemize}
In addition to the starting kit, the important links are:

\begin{itemize}
    \item Competition Site: \url{https://codalab.lisn.upsaclay.fr/competitions/3627}
    \item Forum of the competition: \url{https://codalab.lisn.upsaclay.fr/forums/3627}
    \item Contact email: \url{metalearningchallenge@googlegroups.com}
    \item Meta-learning challenge series website: \url{https://metalearning.chalearn.org/}
    \item Twitter account: \url{https://twitter.com/MetaAlbum}
\end{itemize}

\newpage
\section{Competition Submission API}\label{apd:api}
The participants must overwrite three pre-defined classes: 
\begin{itemize}
    \item \textbf{MetaLearner}: It contains the meta-algorithm logic and only the method \texttt{meta\_fit( meta\_train\_gen, meta\_valid\_gen)} has to be overwritten. In general, a MetaLearner is meta-trained and returns a Learner to be meta-tested. However, it is not mandatory to meta-learn in this method; instead, participants are allowed to return a ``hard-coded'' learning algorithm (Learner).
    
    \item \textbf{Learner}: It encapsulates the logic to learn from a new unseen task. Several methods need to be overwritten:
    \begin{itemize}
        \item \texttt{fit(support\_set)}: Fits the Learner to a new unseen task.
        \item \texttt{save(path)}: Saves the Learner in the specified path.
        \item \texttt{load(path)}: Loads the Learner from the saved file(s).
    \end{itemize}
    In general, a Learner is trained on the support set of a meta-test task and returns a Predictor to be tested on the unlabeled query set of that same task.
    
    \item \textbf{Predictor}: It contains the logic used by the Learner to make predictions once it is fitted. The \texttt{predict(query\_set)} method must be overwritten to receive the unlabeled query set of a task, process it with the fitted Learner, and return the predicted labels.
\end{itemize}

\newpage
\section{Competition Leagues} \label{apd:leagues}
\begin{itemize}
    \item \textbf{Free-style league:} Submit a solution obeying basic challenge rules (pre-trained models allowed).
    \item \textbf{Meta-learning league:} Submit a solution that meta-learns from scratch (no pre-training allowed).
    \item \textbf{New-in-ML league:} Be a participant with less than 10 ML publications, none of which have ever been accepted to the main track of a major conference.
    \item \textbf{Women league:} Special league to encourage women since they rarely enter challenges.
    \item \textbf{Participant of a rarely represented country:} Be a participant of a group that is not in the top 10 most represented countries of \href{https://towardsdatascience.com/kaggle-around-the-world-ccea741b2de2}{Kaggle challenge participants}.
\end{itemize}

The same participant or team can compete in several leagues. For the leagues New-in-ML, Women, and Participant of a rarely represented country, in the case of teams, all team members must fulfill the specific requirements of the league to participate in it. Additionally, in these leagues, the participants are free to use either randomly initialized or pre-trained backbones. 

The total pool prize is 4000 EUR, and it will be evenly distributed among the leagues as follows: 400 EUR for the 1$^{st}$ place, 250 EUR for the 2$^{nd}$ place, and 150 EUR for the 3$^{rd}$ place. Furthermore, we will invite the winning participants to work on a post-challenge analysis collaborative paper.

\newpage

\section{Baselines description and hyperparameters} \label{apd:hyperparameters}
This appendix provides a high-level description of each baseline method with the corresponding hyperparameters used for computing the results presented in this work.

\subsection{Train-from-scratch}
This method is the simplest baseline since it does not perform any meta-training. Therefore, at meta-test time, it trains the backbone with the support set of each unseen task and then uses the trained backbone to predict the labels of the unlabeled query set. Table~\ref{tab:tfs} shows the hyperparameters used by this method.

\subsection{Fine-tuning}
It is a simple transfer learning method that pre-trains a backbone network with batches of data from the concatenated meta-training datasets. Then, at meta-test time, it freezes all the backbone layers except for the last one, which is fine-tuned with the support set of each unseen task. Lastly, the fine-tuned backbone is used to predict the labels of the unlabeled query set of each meta-test task. Table~\ref{tab:finetuning} shows the hyperparameters used by this method.

\subsection{Matching Networks}
In a nutshell, Matching Networks is a metric-based method that performs the following steps:
\begin{enumerate}
    \item Project the images of the support set into the feature space (output embeddings of the backbone). 
    \item Normalize the computed projections of the previous step.
    \item Apply a one-hot encoding on the labels of the support set.
    \item Project the images of the unlabeled query set into the feature space using the same backbone.
    \item Normalize the computed projections of the previous step.
    \item Compute the cosine similarity matrix between the normalized projections of the support and query sets.
    \item Multiply the cosine similarity matrix and the one-hot encoding support set matrix.
    \item Each image in the query set is assigned the label of the column with the highest value in the matrix obtained in the previous step.
\end{enumerate}
During meta-training, the backbone is trained in an``episodic" way by maximizing the cosine similarity between images of the same class in the support and query sets. Please refer to the original paper to see the full details of this method  \cite{matchingnetwork}. Table~\ref{tab:matchingnet} shows the hyperparameters used by this method.

\subsection{Prototypical Networks}
In a nutshell, Prototypical Networks is a metric-based method that performs the following steps:
\begin{enumerate}
    \item Project the images of the support set into the feature space (output embeddings of the backbone). 
    \item Compute the prototypes for each class. The prototypes are the mean vector of all examples of the same class. 
    \item Project the images of the unlabeled query set into the feature space using the same backbone.
    \item Create a distance matrix by computing the Euclidean distance between the projections of the query set and each prototype.
    \item Each image in the query set is assigned the label of the closest prototype.
\end{enumerate}
During meta-training, the backbone is trained in an ``episodic" way by minimizing the Euclidean distance between the projections of the query set images and their corresponding prototypes. Please refer to the original paper to see the full details of this method \cite{Snell2017}. Table~\ref{tab:matchingnet} shows the hyperparameters used by this method.

\subsection{FO-MAML}
In a nutshell, FO-MAML is a method that relies on ``episodic" training to find, during meta-training, the best possible weights for the backbone to achieve rapid adaptation when dealing with unseen tasks (meta-testing). It performs the following steps:
\begin{enumerate}
    \item Forward the support set through the backbone and compute the loss.
    \item Compute the gradients of step 1 and update the weights of the backbone.
    \item Repeat steps 1 and 2 $T$ times.
    \item Forward the query set through the backbone.
    \item If meta-testing, compute the softmax of the output of the previous step and return the probability matrix; otherwise, continue with the next step.
    \item Compute the loss using the output of step 4 and the labels of the query set.
    \item Compute the gradients of step 6, but only update the weights of the backbone after processing $m$ meta-training tasks (meta batch size). 
\end{enumerate}
Note that in the original paper \cite{Finn2017}, MAML computes second-order derivatives during the gradient calculations; nevertheless, this can be avoided by the first-order approximation of MAML (FO-MAML), where the second derivatives are omitted. Table~\ref{tab:maml} shows the hyperparameters used by this method.

\subsection{MetaDelta++}
This baseline corresponds to the solution of the winners of the NeurIPS'21 competition on within-domain few-shot learning. In a nutshell, during meta-training, this method uses batch training to pre-train an ensemble of meta-learners composed of the same backbone but with a different classifier. Additionally, hand-crafted data augmentation (like rotation) is designed to help the pre-training process. Finally, the ensemble of meta-learners is ``autoensembled'' to stabilize the performance of the whole meta-learning system. The ``autoensembled'' backbone is used during meta-testing. Please refer to the original paper to see the full details of this method \cite{metadelta}.

\subsection{Hyperparameters}

\begin{table}[!htpb]
\centering
\caption{Hyperparameters used by the Train-from-scratch baseline. Take into account that the backbone is a ResNet-18.}
\label{tab:tfs}
\begin{tabular}{p{5.3cm}p{3.5cm}p{2.4cm}}
\toprule
Hyperparameter & \begin{tabular}[c]{@{}l@{}}Hyperparameter \\ name in the code\end{tabular} & Value \\
\midrule
Optimizer & \texttt{opt\_fn} & Adam \\
Learning rate & \texttt{lr} & 0.001 \\
Loss function & \texttt{criterion} & Cross-entropy \\
Number of training iterations & \texttt{T} & 100 \\
Batch size & \texttt{batch\_size} & 4 \\
\bottomrule
\end{tabular}
\end{table}

\begin{table}[!htpb]
\centering
\caption{Hyperparameters used by the Fine-tuning baseline. Take into account that the backbone is a ResNet-18.}
\label{tab:finetuning}
\begin{tabular}{p{5.3cm}p{3.5cm}p{2.4cm}}
\toprule
Hyperparameter & \begin{tabular}[c]{@{}l@{}}Hyperparameter \\ name in the code\end{tabular} & Value \\
\midrule
Optimizer & \texttt{opt\_fn} & Adam \\
MetaLearner learning rate & \texttt{lr} & 0.001 \\
Learner learning rate & \texttt{val\_lr} & 0.001 \\
Loss function & \texttt{criterion} & Cross-entropy \\
\begin{tabular}[c]{@{}l@{}}Number of training iterations\\ for the Learner\end{tabular} & \texttt{T} & 100 \\
Batch size for the Learner & \texttt{val\_batch\_size} & 4 \\
\bottomrule
\end{tabular}
\end{table}

\begin{table}[!ht]
\centering
\caption{Hyperparameters used by the Matching Networks and Prototypical Networks baselines. Take into account that the backbone is a ResNet-18.}
\label{tab:matchingnet}
\begin{tabular}{p{3.4cm}p{3.5cm}p{2.4cm}}
\toprule
Hyperparameter & \begin{tabular}[c]{@{}l@{}}Hyperparameter \\ name in the code\end{tabular} & Value \\
\midrule
Optimizer & \texttt{opt\_fn} & Adam \\
Learning rate & \texttt{lr} & 0.001 \\
Loss function & \texttt{criterion} & Cross-entropy \\
Meta-batch size & \texttt{meta\_batch\_size} & 1 \\
\bottomrule
\end{tabular}
\end{table}

\begin{table}[!ht]
\centering
\caption{Hyperparameters used by the FO-MAML baseline. Take into account that the backbone is a ResNet-18.}
\label{tab:maml}
\begin{tabular}{p{5cm}p{3.5cm}p{2.4cm}}
\toprule
Hyperparameter & \begin{tabular}[c]{@{}l@{}}Hyperparameter \\ name in the code\end{tabular} & Value \\
\midrule
Optimizer & \texttt{opt\_fn} & Adam \\
MetaLearner learning rate & \texttt{lr} & 0.001 \\
Learner learning rate & \texttt{base\_lr} & 0.01 \\
Loss function & \texttt{criterion} & Cross-entropy \\
Inner training iterations & \texttt{T} & 5 \\
Meta-batch size & \texttt{meta\_batch\_size} & 2 \\
\bottomrule
\end{tabular}
\end{table}

\clearpage

\section{Detailed results} \label{apd:results}
This appendix provides the detailed results for the experiments shown in \sectionref{sec:results}. Table~\ref{tab:fig1} corresponds to the detailed results for \figureref{fig:baselines}. Tables \ref{tab:fig3-ways-no-pretrained}, \ref{tab:fig3-ways-pretrained}, \ref{tab:fig3-shots-no-pretrained}, and \ref{tab:fig3-shots-pretrained} correspond to the detailed results for \figureref{fig:config-impact}. Lastly, Table~\ref{tab:fig2} corresponds to the detailed results for \figureref{fig:datasets-difficulty}.

\begin{table}[!htpb]
\centering
\caption{Detailed results for the comparison of ``within-domain'' and ``cross-domain'' few-shot learning using a randomly initialized and a pre-trained backbone. The corresponding 95\% CIs are computed at task level over 3,000 tasks.}
\label{tab:fig1}
\begin{tabular}{p{3.7cm}|p{2cm}p{2cm}p{2cm}p{2cm}}
\toprule
\multirow{2}{*}{Method} & \multicolumn{2}{c}{Within-Domain} & \multicolumn{2}{c}{Cross-Domain} \\
 & Random & Pre-trained & Random & Pre-trained \\
 \midrule
Train-from-scratch & 21.1 $\pm$ 0.8 & 33.3 $\pm$ 0.8 & 14.5 $\pm$ 0.7 & 26.9 $\pm$ 0.8 \\
Fine-tuning & 29.0 $\pm$ 0.9 & 39.4 $\pm$ 0.9 & 26.6 $\pm$ 0.8 & 28.8 $\pm$ 0.8 \\
Matching Networks & 35.1 $\pm$ 0.9 & 50.0 $\pm$ 0.9 & 25.7 $\pm$ 0.8 & 30.5 $\pm$ 0.8 \\
Prototypical Networks & 35.2 $\pm$ 0.9 & 51.4 $\pm$ 0.9 & 34.7 $\pm$ 0.9 & 40.6 $\pm$ 0.9 \\
FO-MAML & 24.8 $\pm$ 0.9 & 41.3 $\pm$ 0.9 & 21.1 $\pm$ 0.8 & 21.2 $\pm$ 0.7 \\
MetaDelta++ & 23.6 $\pm$ 0.8 & 75.8 $\pm$ 0.9 & 21.9 $\pm$ 0.9 & 62.3 $\pm$ 1.0 \\
\bottomrule
\end{tabular}
\end{table}

\begin{table}[!htpb]
\centering
\caption{Detailed results for the analysis of the influence of the number of ways on the performance of the baselines in the ``cross-domain'' setting using a randomly initialized backbone. The corresponding 95\% CIs are computed at task level. TFS, FT, MN, PN, and MD++ stands for Train-from-scratch, Fine-tuning, Matching Networks, Prototypical Networks, and MetaDelta++, respectively.}
\label{tab:fig3-ways-no-pretrained}
\begin{tabular}{p{1.1cm}p{1.7cm}p{1.7cm}p{1.7cm}p{1.7cm}p{1.7cm}p{1.7cm}}
\toprule
Ways & TFS & FT & MN & PN & FO-MAML & MD++ \\
 \midrule
2 & 38.0 $\pm$ 5.0 & 49.7 $\pm$ 4.4 & 42.7 $\pm$ 4.9 & 49.8 $\pm$ 4.8 & 38.3 $\pm$ 4.7 & 35.2 $\pm$ 5.6 \\
3 & 29.5 $\pm$ 4.5 & 41.2 $\pm$ 4.6 & 36.9 $\pm$ 4.4 & 45.3 $\pm$ 4.6 & 29.6 $\pm$ 4.4 & 29.3 $\pm$ 4.9 \\
4 & 22.8 $\pm$ 3.5 & 39.9 $\pm$ 4.1 & 34.2 $\pm$ 3.7 & 44.0 $\pm$ 3.9 & 30.0 $\pm$ 3.9 & 28.5 $\pm$ 4.4 \\
5 & 21.0 $\pm$ 3.3 & 35.8 $\pm$ 3.6 & 31.3 $\pm$ 3.3 & 39.4 $\pm$ 3.7 & 26.5 $\pm$ 3.5 & 26.8 $\pm$ 3.7 \\
6 & 18.2 $\pm$ 3.2 & 31.9 $\pm$ 3.7 & 28.6 $\pm$ 3.6 & 38.6 $\pm$ 4.0 & 25.0 $\pm$ 3.5 & 24.4 $\pm$ 4.0 \\
7 & 17.2 $\pm$ 3.1 & 32.8 $\pm$ 3.8 & 28.7 $\pm$ 3.6 & 37.6 $\pm$ 4.0 & 23.9 $\pm$ 3.5 & 24.1 $\pm$ 4.0 \\
8 & 14.5 $\pm$ 2.4 & 27.5 $\pm$ 3.4 & 26.3 $\pm$ 3.2 & 34.3 $\pm$ 3.7 & 22.9 $\pm$ 3.2 & 21.7 $\pm$ 3.7 \\
9 & 15.0 $\pm$ 2.8 & 29.2 $\pm$ 3.6 & 28.0 $\pm$ 3.4 & 37.0 $\pm$ 4.1 & 24.6 $\pm$ 3.5 & 24.4 $\pm$ 4.0 \\
10 & 12.9 $\pm$ 2.5 & 24.4 $\pm$ 3.1 & 23.9 $\pm$ 2.9 & 34.2 $\pm$ 3.7 & 20.4 $\pm$ 3.1 & 20.4 $\pm$ 3.4 \\
11 & 11.0 $\pm$ 2.3 & 23.4 $\pm$ 3.2 & 22.6 $\pm$ 3.1 & 32.0 $\pm$ 3.9 & 18.9 $\pm$ 3.0 & 20.3 $\pm$ 3.5 \\
12 & 10.3 $\pm$ 2.1 & 21.2 $\pm$ 2.8 & 21.7 $\pm$ 2.7 & 31.1 $\pm$ 3.4 & 17.2 $\pm$ 2.5 & 17.6 $\pm$ 2.9 \\
13 & 8.8 $\pm$ 1.8 & 19.4 $\pm$ 2.9 & 20.1 $\pm$ 2.8 & 29.1 $\pm$ 3.5 & 16.3 $\pm$ 2.8 & 17.0 $\pm$ 3.2 \\
14 & 8.6 $\pm$ 1.8 & 20.5 $\pm$ 2.6 & 20.5 $\pm$ 2.7 & 31.2 $\pm$ 3.5 & 16.8 $\pm$ 2.6 & 17.6 $\pm$ 3.1 \\
15 & 7.9 $\pm$ 1.8 & 16.6 $\pm$ 2.5 & 19.4 $\pm$ 2.9 & 26.6 $\pm$ 3.6 & 14.6 $\pm$ 2.6 & 16.2 $\pm$ 3.1 \\
16 & 10.7 $\pm$ 2.2 & 21.8 $\pm$ 2.9 & 23.7 $\pm$ 3.3 & 35.0 $\pm$ 4.1 & 19.0 $\pm$ 3.0 & 21.8 $\pm$ 3.8 \\
17 & 6.0 $\pm$ 1.4 & 16.1 $\pm$ 2.6 & 17.8 $\pm$ 2.7 & 26.7 $\pm$ 3.5 & 13.4 $\pm$ 2.4 & 15.2 $\pm$ 3.0 \\
18 & 6.7 $\pm$ 1.5 & 16.2 $\pm$ 2.5 & 19.2 $\pm$ 2.8 & 27.5 $\pm$ 3.7 & 13.4 $\pm$ 2.5 & 16.9 $\pm$ 3.2 \\
19 & 7.1 $\pm$ 1.7 & 16.2 $\pm$ 2.4 & 18.7 $\pm$ 2.9 & 28.7 $\pm$ 3.8 & 13.3 $\pm$ 2.5 & 17.3 $\pm$ 3.3 \\
20 & 6.3 $\pm$ 1.3 & 17.0 $\pm$ 2.3 & 20.5 $\pm$ 2.8 & 28.5 $\pm$ 3.5 & 13.8 $\pm$ 2.2 & 18.3 $\pm$ 3.1 \\
\bottomrule
\end{tabular}
\end{table}

\begin{table}[!htpb]
\centering
\caption{Detailed results for the analysis of the influence of the number of ways on the performance of the baselines in the ``cross-domain'' setting using a pre-trained backbone. The corresponding 95\% CIs are computed at task level. TFS, FT, MN, PN, and MD++ stands for Train-from-scratch, Fine-tuning, Matching Networks, Prototypical Networks, and MetaDelta++, respectively.}
\label{tab:fig3-ways-pretrained}
\begin{tabular}{p{1.1cm}p{1.7cm}p{1.7cm}p{1.7cm}p{1.7cm}p{1.7cm}p{1.7cm}}
\toprule
Ways & TFS & FT & MN & PN & FO-MAML & MD++ \\
 \midrule
2 & 48.6 $\pm$ 4.4 & 48.5 $\pm$ 4.9 & 50.1 $\pm$ 4.6 & 58.0 $\pm$ 4.3 & 39.2 $\pm$ 4.6 & 76.0 $\pm$ 4.7 \\
3 & 41.4 $\pm$ 4.2 & 43.3 $\pm$ 4.6 & 43.7 $\pm$ 4.5 & 52.8 $\pm$ 4.4 & 32.8 $\pm$ 4.3 & 72.2 $\pm$ 4.7 \\
4 & 39.1 $\pm$ 3.9 & 41.2 $\pm$ 4.0 & 41.5 $\pm$ 3.8 & 50.3 $\pm$ 4.0 & 32.0 $\pm$ 3.7 & 67.1 $\pm$ 4.8 \\
5 & 36.7 $\pm$ 3.7 & 37.1 $\pm$ 3.5 & 37.2 $\pm$ 3.5 & 46.8 $\pm$ 3.6 & 25.7 $\pm$ 3.0 & 67.7 $\pm$ 4.1 \\
6 & 33.1 $\pm$ 3.7 & 35.3 $\pm$ 3.6 & 34.8 $\pm$ 3.7 & 45.1 $\pm$ 4.0 & 25.2 $\pm$ 3.2 & 64.8 $\pm$ 4.5 \\
7 & 32.5 $\pm$ 3.7 & 34.3 $\pm$ 3.7 & 33.9 $\pm$ 3.5 & 43.8 $\pm$ 4.0 & 25.1 $\pm$ 3.2 & 66.8 $\pm$ 4.3 \\
8 & 29.9 $\pm$ 3.3 & 29.1 $\pm$ 3.4 & 30.0 $\pm$ 3.3 & 40.0 $\pm$ 3.9 & 21.4 $\pm$ 3.0 & 63.0 $\pm$ 4.3 \\
9 & 30.6 $\pm$ 3.8 & 30.9 $\pm$ 3.5 & 31.7 $\pm$ 3.4 & 43.2 $\pm$ 4.2 & 22.1 $\pm$ 3.0 & 65.7 $\pm$ 4.3 \\
10 & 26.2 $\pm$ 3.3 & 27.6 $\pm$ 3.1 & 29.5 $\pm$ 3.3 & 39.8 $\pm$ 3.9 & 20.0 $\pm$ 2.8 & 61.3 $\pm$ 4.4 \\
11 & 22.7 $\pm$ 3.2 & 26.4 $\pm$ 3.2 & 28.0 $\pm$ 3.2 & 38.1 $\pm$ 4.0 & 18.9 $\pm$ 2.7 & 61.0 $\pm$ 4.4 \\
12 & 22.4 $\pm$ 3.0 & 23.1 $\pm$ 2.8 & 25.2 $\pm$ 2.8 & 35.2 $\pm$ 3.6 & 17.0 $\pm$ 2.3 & 56.6 $\pm$ 4.1 \\
13 & 20.6 $\pm$ 2.9 & 22.2 $\pm$ 2.9 & 24.4 $\pm$ 3.1 & 34.3 $\pm$ 3.8 & 16.3 $\pm$ 2.5 & 57.9 $\pm$ 4.4 \\
14 & 19.7 $\pm$ 2.6 & 23.7 $\pm$ 2.7 & 26.1 $\pm$ 2.9 & 36.9 $\pm$ 3.7 & 17.7 $\pm$ 2.4 & 58.5 $\pm$ 3.9 \\
15 & 17.9 $\pm$ 2.9 & 18.7 $\pm$ 2.6 & 22.6 $\pm$ 2.9 & 32.3 $\pm$ 3.8 & 14.5 $\pm$ 2.2 & 53.5 $\pm$ 4.6 \\
16 & 22.7 $\pm$ 3.3 & 24.8 $\pm$ 2.9 & 28.4 $\pm$ 3.1 & 40.7 $\pm$ 4.3 & 18.4 $\pm$ 2.5 & 60.8 $\pm$ 4.3 \\
17 & 15.2 $\pm$ 2.4 & 18.7 $\pm$ 2.6 & 21.1 $\pm$ 2.7 & 31.4 $\pm$ 3.6 & 13.0 $\pm$ 2.0 & 54.4 $\pm$ 4.1 \\
18 & 16.0 $\pm$ 2.5 & 18.7 $\pm$ 2.7 & 21.6 $\pm$ 2.8 & 31.8 $\pm$ 3.9 & 13.0 $\pm$ 2.0 & 56.1 $\pm$ 4.3 \\
19 & 15.3 $\pm$ 2.6 & 19.8 $\pm$ 2.6 & 23.6 $\pm$ 3.0 & 34.3 $\pm$ 4.0 & 14.1 $\pm$ 2.1 & 57.9 $\pm$ 4.3 \\
20 & 15.7 $\pm$ 2.2 & 19.5 $\pm$ 2.3 & 23.2 $\pm$ 2.6 & 34.4 $\pm$ 3.7 & 13.3 $\pm$ 1.8 & 59.4 $\pm$ 3.6 \\
\bottomrule
\end{tabular}
\end{table}

\begin{table}[!htpb]
\centering
\caption{Detailed results for the analysis of the influence of the number of shots on the performance of the baselines in the ``cross-domain'' setting using a randomly initialized backbone. The corresponding 95\% CIs are computed at task level. TFS, FT, MN, PN, and MD++ stands for Train-from-scratch, Fine-tuning, Matching Networks, Prototypical Networks, and MetaDelta++, respectively.}
\label{tab:fig3-shots-no-pretrained}
\begin{tabular}{p{1.1cm}p{1.7cm}p{1.7cm}p{1.7cm}p{1.7cm}p{1.7cm}p{1.7cm}}
\toprule
Shots & TFS & FT & MN & PN & FO-MAML & MD++ \\
 \midrule
1 & 8.1 $\pm$ 2.0 & 16.4 $\pm$ 2.9 & 14.2 $\pm$ 2.9 & 16.6 $\pm$ 3.1 & 10.8 $\pm$ 2.2 & 12.7 $\pm$ 3.4 \\
2 & 12.5 $\pm$ 2.5 & 20.7 $\pm$ 3.2 & 18.6 $\pm$ 3.1 & 23.8 $\pm$ 3.5 & 15.6 $\pm$ 2.8 & 16.3 $\pm$ 3.4 \\
3 & 12.8 $\pm$ 2.9 & 24.0 $\pm$ 3.5 & 19.6 $\pm$ 3.3 & 26.3 $\pm$ 3.6 & 17.1 $\pm$ 3.1 & 17.1 $\pm$ 3.6 \\
4 & 13.0 $\pm$ 2.9 & 23.8 $\pm$ 3.4 & 22.0 $\pm$ 3.2 & 30.2 $\pm$ 3.7 & 18.0 $\pm$ 3.1 & 18.2 $\pm$ 3.6 \\
5 & 14.2 $\pm$ 2.6 & 28.2 $\pm$ 3.3 & 24.2 $\pm$ 3.3 & 34.3 $\pm$ 3.8 & 20.9 $\pm$ 3.0 & 20.6 $\pm$ 3.4 \\
6 & 15.0 $\pm$ 3.1 & 25.8 $\pm$ 3.6 & 25.0 $\pm$ 3.5 & 32.8 $\pm$ 4.0 & 21.1 $\pm$ 3.4 & 20.0 $\pm$ 3.8 \\
7 & 14.1 $\pm$ 3.3 & 25.5 $\pm$ 3.8 & 24.4 $\pm$ 3.6 & 33.4 $\pm$ 4.2 & 19.7 $\pm$ 3.4 & 20.6 $\pm$ 4.0 \\
8 & 15.8 $\pm$ 3.3 & 28.7 $\pm$ 3.9 & 26.9 $\pm$ 3.7 & 36.8 $\pm$ 4.2 & 22.5 $\pm$ 3.7 & 22.9 $\pm$ 3.9 \\
9 & 16.4 $\pm$ 3.8 & 28.4 $\pm$ 4.3 & 26.6 $\pm$ 4.1 & 37.0 $\pm$ 4.5 & 23.3 $\pm$ 4.1 & 23.4 $\pm$ 4.5 \\
10 & 14.7 $\pm$ 3.1 & 27.0 $\pm$ 3.6 & 27.5 $\pm$ 3.2 & 36.4 $\pm$ 3.9 & 21.2 $\pm$ 3.4 & 23.9 $\pm$ 3.7 \\
11 & 15.9 $\pm$ 3.1 & 29.7 $\pm$ 3.5 & 29.1 $\pm$ 3.5 & 38.7 $\pm$ 3.9 & 24.4 $\pm$ 3.6 & 24.1 $\pm$ 4.0 \\
12 & 14.4 $\pm$ 3.0 & 27.3 $\pm$ 3.7 & 26.8 $\pm$ 3.5 & 36.6 $\pm$ 3.9 & 22.2 $\pm$ 3.5 & 23.0 $\pm$ 4.0 \\
13 & 15.3 $\pm$ 3.0 & 28.3 $\pm$ 3.8 & 28.7 $\pm$ 3.5 & 38.7 $\pm$ 4.0 & 22.9 $\pm$ 3.4 & 25.0 $\pm$ 3.9 \\
14 & 15.7 $\pm$ 4.1 & 28.2 $\pm$ 4.3 & 28.5 $\pm$ 4.2 & 39.5 $\pm$ 4.7 & 23.5 $\pm$ 4.1 & 24.2 $\pm$ 4.6 \\
15 & 15.2 $\pm$ 3.3 & 28.2 $\pm$ 4.0 & 28.0 $\pm$ 3.7 & 38.6 $\pm$ 4.6 & 22.3 $\pm$ 3.6 & 25.6 $\pm$ 4.2 \\
16 & 18.3 $\pm$ 4.1 & 30.8 $\pm$ 4.4 & 30.7 $\pm$ 4.3 & 41.6 $\pm$ 4.3 & 24.5 $\pm$ 4.2 & 25.8 $\pm$ 4.8 \\
17 & 14.1 $\pm$ 3.5 & 28.6 $\pm$ 4.0 & 27.6 $\pm$ 4.0 & 38.3 $\pm$ 4.4 & 22.7 $\pm$ 3.9 & 24.1 $\pm$ 4.4 \\
18 & 13.6 $\pm$ 2.5 & 26.5 $\pm$ 3.1 & 28.1 $\pm$ 2.9 & 38.6 $\pm$ 3.5 & 22.8 $\pm$ 3.0 & 23.2 $\pm$ 3.3 \\
19 & 16.5 $\pm$ 3.9 & 29.4 $\pm$ 4.2 & 31.3 $\pm$ 3.7 & 40.8 $\pm$ 4.1 & 26.2 $\pm$ 4.0 & 25.8 $\pm$ 4.3 \\
20 & 16.6 $\pm$ 3.4 & 29.0 $\pm$ 3.9 & 29.2 $\pm$ 3.4 & 40.6 $\pm$ 4.0 & 23.3 $\pm$ 3.6 & 24.2 $\pm$ 4.1 \\
\bottomrule
\end{tabular}
\end{table}

\begin{table}[!htpb]
\centering
\caption{Detailed results for the analysis of the influence of the number of shots on the performance of the baselines in the ``cross-domain'' setting using a pre-trained backbone. The corresponding 95\% CIs are computed at task level. TFS, FT, MN, PN, and MD++ stands for Train-from-scratch, Fine-tuning, Matching Networks, Prototypical Networks, and MetaDelta++, respectively.}
\label{tab:fig3-shots-pretrained}
\begin{tabular}{p{1.1cm}p{1.7cm}p{1.7cm}p{1.7cm}p{1.7cm}p{1.7cm}p{1.7cm}}
\toprule
Shots & TFS & FT & MN & PN & FO-MAML & MD++ \\
 \midrule
1 & 16.0 $\pm$ 2.8 & 16.9 $\pm$ 3.0 & 18.1 $\pm$ 3.2 & 23.7 $\pm$ 3.5 & 11.0 $\pm$ 2.3 & 48.5 $\pm$ 4.8 \\
2 & 21.5 $\pm$ 3.2 & 22.0 $\pm$ 3.4 & 24.0 $\pm$ 3.4 & 29.8 $\pm$ 4.1 & 14.9 $\pm$ 2.4 & 50.6 $\pm$ 4.7 \\
3 & 22.9 $\pm$ 3.2 & 24.2 $\pm$ 3.5 & 25.7 $\pm$ 3.6 & 32.2 $\pm$ 3.8 & 17.4 $\pm$ 2.8 & 55.3 $\pm$ 4.6 \\
4 & 23.5 $\pm$ 3.4 & 25.0 $\pm$ 3.5 & 26.4 $\pm$ 3.4 & 35.4 $\pm$ 3.9 & 18.5 $\pm$ 2.7 & 58.4 $\pm$ 4.6 \\
5 & 26.3 $\pm$ 3.1 & 29.7 $\pm$ 3.3 & 30.6 $\pm$ 3.4 & 41.1 $\pm$ 3.9 & 21.0 $\pm$ 2.7 & 62.7 $\pm$ 4.3 \\
6 & 26.6 $\pm$ 3.4 & 27.6 $\pm$ 3.5 & 29.6 $\pm$ 3.6 & 38.7 $\pm$ 4.1 & 19.8 $\pm$ 3.1 & 59.5 $\pm$ 4.4 \\
7 & 27.4 $\pm$ 3.7 & 28.4 $\pm$ 3.8 & 29.4 $\pm$ 3.7 & 39.5 $\pm$ 4.2 & 20.4 $\pm$ 3.2 & 62.8 $\pm$ 4.6 \\
8 & 27.9 $\pm$ 3.8 & 30.7 $\pm$ 3.7 & 32.9 $\pm$ 3.6 & 43.2 $\pm$ 4.2 & 23.3 $\pm$ 3.2 & 60.7 $\pm$ 4.4 \\
9 & 28.9 $\pm$ 4.2 & 32.3 $\pm$ 4.1 & 34.6 $\pm$ 4.2 & 43.6 $\pm$ 4.6 & 25.6 $\pm$ 4.1 & 63.0 $\pm$ 5.1 \\
10 & 28.5 $\pm$ 3.7 & 29.9 $\pm$ 3.5 & 30.8 $\pm$ 3.5 & 43.5 $\pm$ 4.1 & 21.1 $\pm$ 3.0 & 65.2 $\pm$ 4.3 \\
11 & 30.6 $\pm$ 3.7 & 32.9 $\pm$ 3.5 & 34.3 $\pm$ 3.6 & 45.3 $\pm$ 4.0 & 24.6 $\pm$ 3.4 & 66.6 $\pm$ 4.1 \\
12 & 26.2 $\pm$ 3.6 & 30.3 $\pm$ 3.7 & 32.8 $\pm$ 3.6 & 43.3 $\pm$ 3.9 & 23.6 $\pm$ 3.4 & 63.6 $\pm$ 4.4 \\
13 & 30.3 $\pm$ 3.8 & 30.2 $\pm$ 3.6 & 32.1 $\pm$ 3.5 & 44.4 $\pm$ 4.0 & 23.1 $\pm$ 3.4 & 67.6 $\pm$ 4.1 \\
14 & 29.7 $\pm$ 4.6 & 31.0 $\pm$ 4.2 & 32.6 $\pm$ 4.1 & 44.9 $\pm$ 4.6 & 23.0 $\pm$ 3.6 & 65.8 $\pm$ 4.8 \\
15 & 28.6 $\pm$ 3.9 & 30.6 $\pm$ 4.0 & 33.0 $\pm$ 3.7 & 44.1 $\pm$ 4.5 & 22.8 $\pm$ 3.4 & 64.7 $\pm$ 4.5 \\
16 & 29.5 $\pm$ 4.4 & 34.5 $\pm$ 4.4 & 36.0 $\pm$ 4.1 & 47.1 $\pm$ 4.6 & 25.6 $\pm$ 4.0 & 69.2 $\pm$ 4.3 \\
17 & 28.4 $\pm$ 4.3 & 30.9 $\pm$ 4.0 & 33.9 $\pm$ 4.1 & 44.4 $\pm$ 4.5 & 24.0 $\pm$ 3.8 & 65.5 $\pm$ 4.3 \\
18 & 28.0 $\pm$ 3.5 & 29.5 $\pm$ 3.2 & 31.0 $\pm$ 3.0 & 43.4 $\pm$ 3.5 & 21.2 $\pm$ 2.6 & 66.4 $\pm$ 3.8 \\
19 & 30.2 $\pm$ 4.3 & 31.5 $\pm$ 4.3 & 34.1 $\pm$ 3.9 & 45.2 $\pm$ 4.3 & 23.6 $\pm$ 3.7 & 67.0 $\pm$ 4.4 \\
20 & 29.4 $\pm$ 4.1 & 31.0 $\pm$ 3.8 & 33.0 $\pm$ 3.6 & 45.4 $\pm$ 4.1 & 23.0 $\pm$ 3.3 & 66.5 $\pm$ 4.4 \\
\bottomrule
\end{tabular}
\end{table}

\begin{table}[!htpb]
\centering
\caption{Detailed results for the comparison of the difficulty level of Feedback phase datasets for ``within-domain'' and ``cross-domain'' few-shot learning. The results of Train-from-scratch use a randomly initialized backbone, while the results of MetaDelta++ use a pre-trained backbone. The corresponding 95\% CIs are computed at task level over 300 tasks.}
\label{tab:fig2}
\begin{tabular}{p{2cm}|p{3.2cm}p{2.7cm}p{3.2cm}p{2.5cm}}
\toprule
\multirow{2}{*}{Dataset} & \multicolumn{2}{c}{Within-Domain} & \multicolumn{2}{c}{Cross-Domain} \\
 & Train-from-scratch & MetaDelta++ & Train-from-scratch & MetaDelta++ \\
 \midrule
Dataset 1 & 9.5 $\pm$ 0.8 & 94.0 $\pm$ 0.7 & 7.9 $\pm$ 1.1 & 87.3 $\pm$ 1.1 \\
Dataset 2 & 9.4 $\pm$ 0.8 & 57.0 $\pm$ 1.6 & 7.0 $\pm$ 1.0 & 45.7 $\pm$ 1.7 \\
Dataset 3 & 16.2 $\pm$ 1.0 & 58.4 $\pm$ 1.4 & 9.4 $\pm$ 1.2 & 50.7 $\pm$ 1.7 \\
Dataset 4 & 61.6 $\pm$ 1.4 & 92.6 $\pm$ 1.0 & 50.2 $\pm$ 2.1 & 93.8 $\pm$ 0.6 \\
Dataset 5 & 6.8 $\pm$ 0.7 & 17.3 $\pm$ 0.9 & 5.2 $\pm$ 0.9 & 13.3 $\pm$ 1.0 \\
Dataset 6 & 48.9 $\pm$ 1.4 & 94.5 $\pm$ 0.7 & 31.9 $\pm$ 2.7 & 88.9 $\pm$ 1.0 \\
Dataset 7 & 34.4 $\pm$ 1.3 & 81.8 $\pm$ 0.9 & 18.4 $\pm$ 1.8 & 52.8 $\pm$ 1.7 \\
Dataset 8 & 10.3 $\pm$ 0.8 & 81.6 $\pm$ 0.8 & 6.7 $\pm$ 1.0 & 70.5 $\pm$ 1.4 \\
Dataset 9 & 13.2 $\pm$ 0.9 & 86.7 $\pm$ 1.2 & 7.9 $\pm$ 1.3 & 73.8 $\pm$ 1.5 \\
Dataset 10 & 0.5 $\pm$ 0.5 & 93.7 $\pm$ 0.6 & 0.7 $\pm$ 0.6 & 45.7 $\pm$ 3.4 \\
\bottomrule
\end{tabular}
\end{table}

\end{document}